\documentclass[10pt,twocolumn,letterpaper]{article}

\usepackage{iccv}
\usepackage{times}
\usepackage{epsfig}
\usepackage{graphicx}
\usepackage{amsmath}
\usepackage{amssymb}
\usepackage{multirow}
\usepackage{booktabs}
\usepackage{bbding}
\usepackage{caption}
\usepackage{subcaption}
\usepackage{capt-of}
\usepackage{makecell}
\usepackage{algorithmic}
\usepackage[linesnumbered,ruled,vlined]{algorithm2e}

\newcommand\blfootnote[1]{
    \begingroup
    \renewcommand\thefootnote{}\footnote{#1}
    \addtocounter{footnote}{-1}
    \endgroup
}

\SetKwInput{KwInput}{Input}                % Set the Input
\SetKwInput{KwOutput}{Output}              % set the Output

\SetCommentSty{mycommfont}

\usepackage[pagebackref=true,breaklinks=true,letterpaper=true,colorlinks,bookmarks=false]{hyperref}
\usepackage{lipsum}

\iccvfinalcopy % *** Uncomment this line for the final submission

\def\iccvPaperID{3435} % *** Enter the ICCV Paper ID here

\ificcvfinal\fi

\begin{document}
\definecolor{ForestGreen}{RGB}{34,139,34}
\definecolor{BrightGreen}{RGB}{68,153,69}
\definecolor{DarkBlue}{RGB}{31,112,169}
\definecolor{Orange}{RGB}{234,120,39}

%%%%%%%%% TITLE
\title{SparseFusion: Fusing Multi-Modal Sparse Representations \\
for Multi-Sensor 3D Object Detection}

\author{Yichen Xie$^{1,*}$, Chenfeng Xu$^{1,*}$, Marie-Julie Rakotosaona$^{2}$, Patrick Rim$^{3}$, Federico Tombari$^{2}$,\\ Kurt Keutzer$^{1}$, Masayoshi Tomizuka$^{1}$, Wei Zhan$^{1}$\\ % Author(s)
$^{1}$ University of California, Berkeley $^{2}$ Google $^{3}$ California Institute of Technology\\ % Institution(s)
% {\tt\small \{yichen\_xie,xuchenfeng,tomizuka,wzhan\}@berkeley.edu
}

\maketitle
\ificcvfinal\thispagestyle{empty}\fi
\blfootnote{* indicates equal contribution.}

%%%%%%%%% ABSTRACT
\begin{abstract}
By identifying four important components of existing LiDAR-camera 3D object detection methods (LiDAR and camera candidates, transformation, and fusion outputs), we observe that all existing methods either find dense candidates or yield dense representations of scenes. However, given that objects occupy only a small part of a scene, finding dense candidates and generating dense representations is noisy and inefficient. We propose SparseFusion, a novel multi-sensor 3D detection method that exclusively uses sparse candidates and sparse representations. Specifically, SparseFusion utilizes the outputs of parallel detectors in the LiDAR and camera modalities as sparse candidates for fusion. We transform the camera candidates into the LiDAR coordinate space by disentangling the object representations. Then, we can fuse the multi-modality candidates in a unified 3D space by a lightweight self-attention module. To mitigate negative transfer between modalities, we propose novel semantic and geometric cross-modality transfer modules that are applied prior to the modality-specific detectors. SparseFusion achieves state-of-the-art performance on the nuScenes benchmark while also running at the fastest speed, even outperforming methods with stronger backbones. We perform extensive experiments to demonstrate the effectiveness and efficiency of our modules and overall method pipeline. Our code will be made publicly available at \href{https://github.com/yichen928/SparseFusion}{https://github.com/yichen928/SparseFusion}.
% Besides, we provide sufficient analysis and ablation studies to shed light on how it works.

\end{abstract}
\vspace{-0.5cm}
%%%%%%%%% BODY TEXT
\section{Introduction}
% Camera-LiDAR sensor fusion advances state-of-the-arts in 3D perception of current autonomous driving cars. The two sensors are complementary counterparts, i.e., LiDAR tells the occupancy of the 3D space and Camera provides rich semantic cues.
% However, the significant discrepancy in the modalities of two sensors brings up inevitable challenges for the multi-sensor fusion algorithms.

% Although both point cloud and images are representations of the physical world, point cloud are a set of unordered xyz points, capturing the world in 

Autonomous driving cars rely on multiple sensors, such as LiDAR and cameras, to perceive the surrounding environment. LiDAR sensors provide accurate 3D scene occupancy information through point clouds with points in the xyz coordinate space, and cameras provide rich semantic information through images with pixels in the RGB color space. However, there are often significant discrepancies between representations of the same physical scene acquired by the two sensors, as LiDAR sensors capture point clouds using 360-degree rotation while cameras capture images from a perspective view without a sense of depth. This impedes an effective and efficient fusion of the LiDAR and camera modalities. To tackle this challenge, multi-sensor fusion algorithms were proposed to \textit{find correspondences between multi-modality data to transform and fuse them into a unified scene representation space.}

%现在的主流的scene representation的方式是dense representation like point/BEV/volumes, specifically，通过在将2D image feature和point cloud feature transform到同一个space 然后fuse起来。这种represent 试图去represent 整个3D 空间，这对于3D detection是reduandent（因为我们其实只care instances 并且instances 只占空间的很少一部分）。另一方面，去build这种representaiton 也是不efficient的, 比如将image转成bev 或者volume 是需要非常多的计算量的。

% 一种更efficient的并且也是目前state of the art的方式是sparse representation，通过用query 去表征 object 然后去interact 不同的modality，然而这些query iteratively的和两个完全不同modality之间做interaction，很难保证它能被otimize到最优，尽管current stateofthe art deep interaction 去mitigate domain gap via featureinteraction，但这种feature 之间的interact非常time consuming。我们skip 这个domain gap 以一种非常efficient的方式，makeuseof offtheshelf 2d和3d dectior 我们选取

\begin{figure}[tb!]
    \centering
    \begin{subfigure}{0.48\linewidth}
        \includegraphics[width=\linewidth]{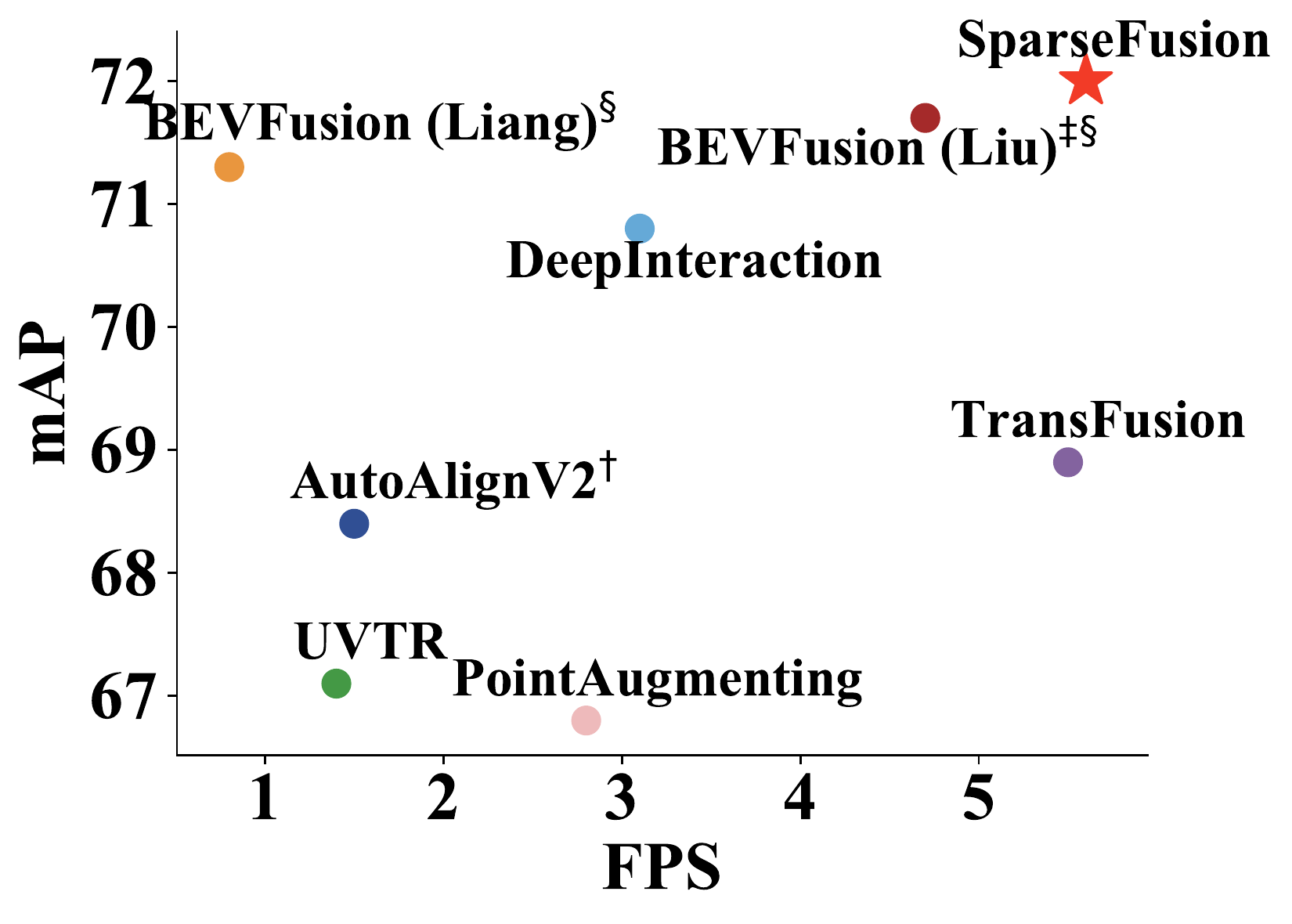}
        \caption{mAP \textit{vs.} FPS}
        \label{fig:trade-off}
    \end{subfigure}
    \hfill
    \begin{subfigure}{0.48\linewidth}
        \includegraphics[width=\linewidth, trim={0 -0.6cm 0 0}]{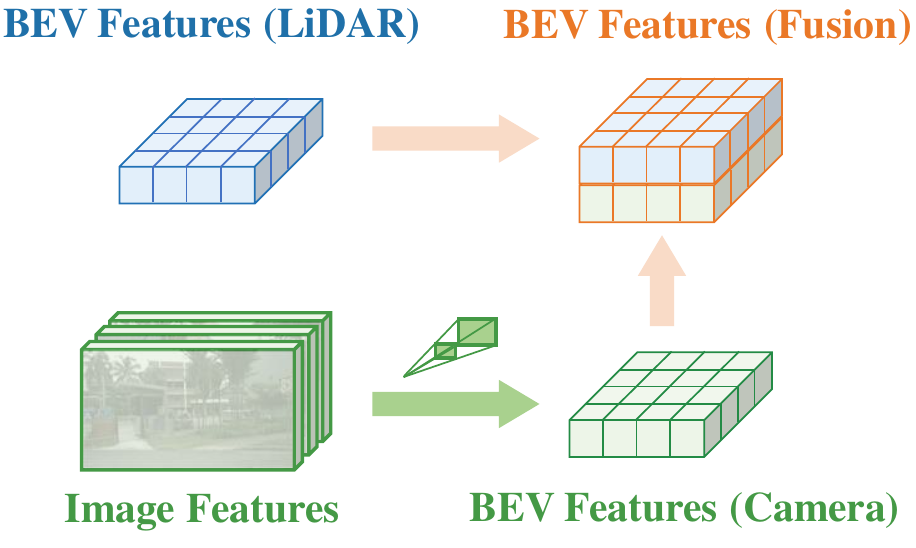}
        \caption{Dense-to-dense fusion.}
        \label{fig:dense_representation}
    \end{subfigure}
    \par\medskip
    \begin{subfigure}{\linewidth}
        \includegraphics[width=\linewidth]{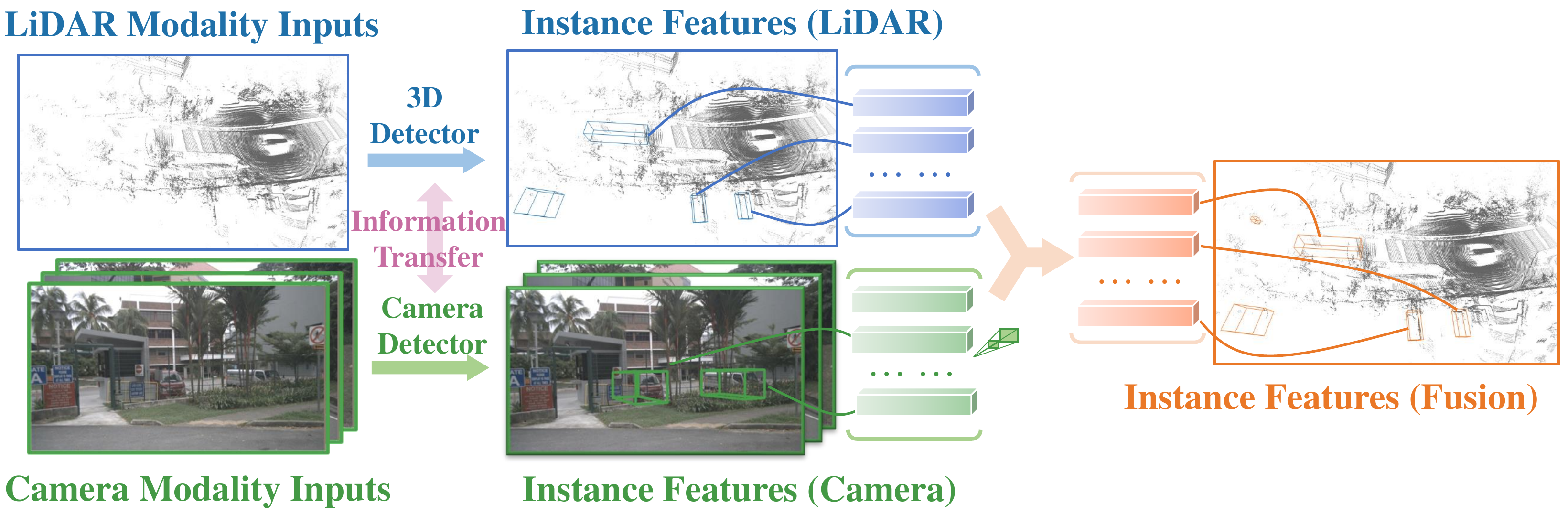}
        \caption{Overview of our sparse fusion strategy. We extract instance-level features from the LiDAR and camera modalities separately, and fuse them in a unified 3D space to perform detection.}
        \label{fig:teaser}
    \end{subfigure}
    \caption{Compared to existing fusion algorithms, SparseFusion achieves state-of-the-art performance as well as the fastest inference speed on nuScenes \textit{test set}. $\dag:$ Official code of \cite{chen2022autoalignv2} uses flip as test-time augmentation. $\ddag:$ We use BEVFusion-base results in the official repository of \cite{liu2022bevfusion} to match the input resolutions of other methods. $\S:$ Swin-T~\cite{liu2021swin,liang2022cbnet} is adopted as image backbone.}
    \vspace{-10pt}
\end{figure}

\begin{table*}[tb!]
    \centering
    \caption{
For each LiDAR-camera fusion method, we identify the \textit{LiDAR candidates} and \textit{camera candidates} that are used, the \textit{transformation} process used to fuse these candidates into a unified space, and the \textit{fusion outputs} generated to represent 3D scenes using information from both modalities. Based on these components, we categorize the methods into the following categories: \textbf{Dense+Sparse$\rightarrow$Dense} approaches relate the sparse region proposals in images to a dense frustum point cloud and fuse them into a dense point space. \textbf{Dense+Dense$\rightarrow$Dense} approaches align each point feature in the point cloud to the corresponding pixel feature in the image and represent the 3D scenes using dense features such as point/BEV features. \textbf{Sparse+Dense$\rightarrow$Sparse} approaches generate sparse queries by detecting instance features in the point cloud and then applying cross-attention with dense image features. \textbf{Dense+Dense$\rightarrow$Sparse} approaches predict objects using queries that combine dense features from each modality. \textbf{Sparse+Sparse$\rightarrow$Sparse (ours)} extracts sparse instance features from each modality and directly fuses them to obtain the final sparse instance features used for detection.
}
    \label{tab:correspondence}
    \vspace{-5pt}

    \resizebox{0.95\linewidth}{!}{
    \begin{tabular}{cccccc}
       \toprule
       \textbf{Category} & \textbf{Method} & \textbf{LiDAR Candidate} & \textbf{Camera Candidate} & \textbf{Transformation} & \textbf{Fusion Outputs} \\
       \midrule
         Dense+Sparse$\rightarrow$Dense & Frustum PointNets~\cite{qi2018frustum} & point features & region proposals & proj. \& concat. & point features\\
       \midrule
         \multirow{5}{*}{Dense+Dense$\rightarrow$Dense} & PointPainting~\cite{vora2020pointpainting} & point features & segm. output & proj. \& concat. & point features\\
         & PointAugmenting~\cite{wang2021pointaugmenting} & point features & image features & proj. \& concat. & point features\\
         & PAI3D~\cite{liu2023pai3d} & point features & segm. output & proj. \& concat. & point features\\
         & BEVFusion~\cite{liu2022bevfusion} & BEV features & image features &  depth. est. \& proj. \& concat. & BEV features\\
         & AutoAlignV2~\cite{chen2022autoalignv2} & voxel features & image features & proj. \& attn.  & voxel features\\
          % & MSMDFusion~\cite{jiao2022msmdfusion} & voxel features & image features &  proj. \& concat. & BEV features\\
        \midrule
        Sparse+Dense$\rightarrow$Sparse & TransFusion~\cite{bai2022transfusion} & instance features & image features & proj. \& attn. & instance features\\
        \midrule
         \multirow{4}{*}{Dense+Dense$\rightarrow$Sparse} & FUTR3D~\cite{chen2022futr3d} & voxel features & image features & attn. & instance features\\
         & UVTR~\cite{li2022unifying} & voxel features & image features & depth. est. \& proj. \& attn. & instance features\\
         & DeepInteraction~\cite{yang2022deepinteraction} & BEV features & image features & proj. \& attn. & instance features\\
         & CMT~\cite{yan2023cross} & BEV features & image features & attn. & instance features\\
        \midrule
          Sparse+Sparse$\rightarrow$Sparse & SparseFusion (\textbf{Ours}) & instance features & instance features & proj. \& attn. & instance features\\
       \bottomrule
    \end{tabular}
    }
\end{table*}

Dense representations, such as bird-eye-view (BEV), volumetric, and point representations, are commonly used to represent 3D scenes \cite{liu2022bevfusion,chen2022autoalignv2,vora2020pointpainting,liu2023pai3d,wang2021pointaugmenting,jiao2022msmdfusion}. Most previous works fuse different modalities by aligning low-level data or high-level features to yield dense features that describe the entire 3D scene, \textit{e.g.}, as shown in Fig~\ref{fig:dense_representation}. However, for the task of 3D object detection, such dense representations are superfluous since we are only interested in instances/objects, which only occupy a small part of the 3D space. Furthermore, noisy backgrounds can be detrimental to object detection performance, and aligning different modalities into the same space is a time-consuming process. For example, generating BEV features from multi-view images takes 500ms on an RTX 3090 GPU \cite{liu2022bevfusion}.

In contrast, sparse representations are more efficient, and methods based on them have achieved state-of-the-art performance in multi-sensor 3D detection \cite{bai2022transfusion,chen2022futr3d,li2022unifying,yang2022deepinteraction}. These methods use object queries to represent instances/objects in the scene and interact with the original image and point cloud features. However, most previous works do not take into account the significant domain gap between features from different modalities \cite{xu2022image2point}. The queries may gather information from one modality that has a large distribution shift with respect to another modality, making iterative interaction between modalities with large gaps sub-optimal. Recent work \cite{yang2022deepinteraction} mitigates this issue by incorporating modality interaction, \textit{i.e.} performing cross-attention between features from two different modalities. However, the number of computations performed in this method increases quadratically with the dimensions of features and is thus inefficient. We categorize previous works into four groups by identifying four key components, which are outlined in Table~\ref{tab:correspondence}. Further discussion of the methods in these groups is presented in Sec. \ref{related_work}.

In this paper, we propose SparseFusion, a novel method (Fig.~\ref{fig:teaser}) that simultaneously utilizes sparse candidates and yields sparse representations, enabling efficient and effective 3D object detection. SparseFusion is the first LiDAR-camera fusion method, to our knowledge, to perform 3D detection using exclusively sparse candidates and sparse fusion outputs.
% overcome the discrepancy between the two modalities. 
We highlight a key common ground between the two modalities: an image and a point cloud that represent the same 3D scene will contain mostly the same instances/objects. To leverage this commonality, we perform 3D object detection on the inputs from each modality in two parallel branches. Then, the instance features from each branch are projected into a unified 3D space. Since the instance-level features are sparse representations of the same objects in the same scene, we are able to fuse them with a lightweight attention module~\cite{vaswani2017attention} in a \textit{soft} manner. This parallel detection strategy allows the LiDAR and camera branches to take advantage of the unique strengths of the point cloud and image representations, respectively.
% while making up for the ignorance of each other. 
Nevertheless, the drawbacks of each single-modality detector may result in negative transfer during the fusion phase. For example, the point cloud detector may struggle to distinguish between a standing person and a tree trunk due to a lack of detailed semantic information, while the image detector is hard to localize objects in the 3D space due to a lack of accurate depth information. To mitigate the issue of negative transfer, we introduce a novel cross-modality information transfer method designed to compensate for the deficiencies of each modality. This method is applied to the inputs from both modalities prior to the parallel detection branches.

% However, the performance of each single-modality detection is hindered by its own shortcomings, \textit{i.e.}, LiDAR modality lacks semantics for classification, and camera modality requires reliable geometry to localize objects \cite{park2022detmatch}. As a result, we enable cross-modality information transfer in the early stage to make up for the per-modality deficiencies. To be specific, we transfer semantic information from the camera to LiDAR through image-guided query initialization to boost the object classification in the LiDAR branch. Simultaneously, the detector in the camera branch borrows the geometric information from LiDAR through a sparse depth map to localize objects in the 3D scene. 

SparseFusion achieves state-of-the-art results on the competitive nuScenes benchmark~\cite{caesar2020nuscenes}. Our instance-level sparse fusion strategy allows for a lighter network and much higher efficiency in comparison with prior work. With the same backbone, SparseFusion outperforms the current state-of-the-art model~\cite{yang2022deepinteraction} with 1.8x acceleration. Our contributions are summarized as follows:
\begin{itemize}
\vspace{-0.2cm}
    \item We revisit prior LiDAR-camera fusion works and identify four important components that allow us to categorize existing methods into four groups. We propose an entirely new category of methods that exclusively uses sparse candidates and representations. 
    % analyze the performance gain of existing LiDAR-camera fusion methods and empirically demonstrate the role of camera inputs in the fusion process. 
    % This challenges the necessity of the popular dense fusion strategy and inspires us to propose a sparse alternative.
\vspace{-0.2cm}
    \item We propose SparseFusion, a novel method for LiDAR-camera 3D object detection that leverages instance-level sparse feature fusion and cross-modality information transfer to take advantage of the strengths of each modality while mitigating their weaknesses.
\vspace{-0.2cm} 
    \item We demonstrate that our method achieves state-of-the-art performance in 3D object detection with a lightweight architecture that provides the fastest inference speed. 
\end{itemize}

\section{Related Work}
\label{related_work}
\noindent\textbf{LiDAR-based 3D Object Detection.} LiDAR sensors are commonly used for single-modality 3D object detection due to the accurate geometric information provided by point clouds. For detection in outdoor scenes, most existing methods transform unordered point clouds into more structured data formats such as pillars~\cite{lang2019pointpillars}, voxels~\cite{yan2018second,deng2021voxel}, or range views~\cite{fan2021rangedet,liang2020rangercnn}. Features are extracted by standard 2D or 3D convolutional networks, based on which a detection head is used to recognize objects and regress 3D bounding boxes. Mainstream detection heads apply anchor-based~\cite{zhou2018voxelnet,lang2019pointpillars} or center-based~\cite{zhou2019objects} structures. Inspired by the promising performance of transformer-based methods in 2D detection, some recent works explore transformers as feature extractors~\cite{mao2021voxel,sheng2021improving} or as detection heads~\cite{bai2022transfusion,misra2021end}. Our method is agnostic to the LiDAR-based detector used in the LiDAR branch, and the default setting uses TransFusion-L~\cite{bai2022transfusion}.
%and the default setting of SparseFusion uses TransFusion-L~\cite{bai2022transfusion} with VoxelNet backbone~\cite{zhou2018voxelnet} as the LiDAR-modality detector. 

\noindent\textbf{Camera-based 3D Object Detection.} Camera-based 3D detection methods are also being studied increasingly. Early work performs monocular 3D object detection by attaching extra 3D bounding box regression heads~\cite{yin2021center,wang2021fcos3d} to 2D detectors. In practice, scenes are often perceived by multiple cameras from different perspective views. Following LSS~\cite{philion2020lift}, methods like BEVDet~\cite{huang2021bevdet} and BEVDepth~\cite{li2022bevdepth} extract 2D features from multi-view images and project them into the BEV space. Other methods including DETR3D~\cite{wang2022detr3d} and PETR~\cite{liu2022petr} adapt techniques from transformer-based 2D object detection methods~\cite{zhu2020deformable,carion2020end} to learn correspondences between different perspective views through cross-attention using 3D queries. However, as revealed in~\cite{huang2022obmo}, there inevitably exists some ambiguity when recovering 3D geometry from 2D images. In response, recent works~\cite{huang2022bevdet4d,li2022bevformer,park2022time} also explore the positive effects of temporal cues in camera-based 3D detection. In our proposed SparseFusion, we extend deformable-DETR~\cite{zhu2020deformable} to monocular 3D object detection and explicitly transform the regressed bounding boxes to the LiDAR coordinate space.

\noindent\textbf{Multi-Modality 3D Object Detection.} LiDAR and cameras provide complementary information about the surrounding environment, so it is appealing to fuse the multi-modality inputs for 3D object detection tasks. As analyzed in Tab.~\ref{tab:correspondence}, existing fusion methods can be classified into four categories. Early works tend to fuse multi-modality information into a unified dense representation. Frustum PointNets~\cite{qi2018frustum} utilizes a \textbf{Dense+Sparse$\rightarrow$Dense} approach that filters dense point clouds with sparse 2D regions of interest. Subsequent works explore \textbf{Dense+Dense$\rightarrow$Dense} approaches by working directly with the dense LiDAR modality and camera modality features instead. Methods such as \cite{vora2020pointpainting,wang2021pointaugmenting} project point clouds into image perspective views and concatenate the dense image features with point features. BEVFusion~\cite{liu2022bevfusion,liang2022bevfusion} significantly improves the performance of this line of methods by projecting dense image features into the LiDAR coordinate space using estimated per-pixel depths. AutoAlignV2~\cite{chen2022autoalignv2} also considers the soft correspondence through cross-modality attention to increase the robustness. However, we point out that dense representations are altogether undesirable for 3D object detection as they are noisy and inefficient. 

Recent works have begun to utilize object-centric sparse scene representations. TransFusion~\cite{bai2022transfusion} adopts a \textbf{Sparse+Dense$\rightarrow$Sparse} strategy by extracting sparse instance-level features from the LiDAR modality and refining them using dense image features. Other works~\cite{chen2022futr3d,li2022unifying,yang2022deepinteraction,yan2023cross} utilize a \textbf{Dense+Dense$\rightarrow$Sparse} approach where queries are used to extract a sparse instance-level representation from dense BEV and image features. However, it is hard to extract information from multi-modality features with an attention operation given the large cross-modal distribution shift. To this end, UVTR~\cite{li2022unifying} projects image features into the LiDAR coordinate space, CMT~\cite{yan2023cross} encodes modality-specific positional information to its queries, and DeepInteraction~\cite{yang2022deepinteraction} proposes cross-modality interaction. However, these methods still need to resolve the large multi-modal domain gap by stacking many transformer layers to construct a heavy decoder.

In contrast to the above methods, our method adopts the previously unexplored \textbf{Sparse+Sparse$\rightarrow$ Sparse} approach. SparseFusion extracts sparse representations of both modalities and fuses them to generate a more accurate and semantically rich sparse representation that yields great performance while also achieving great efficiency.

%explore a sparse-to-sparse instance-level fusion method to exploit the strengths of each modality, achieving a great trade-off between effectiveness and efficiency.

\begin{figure*}
    \centering
    \includegraphics[width=0.85\linewidth]{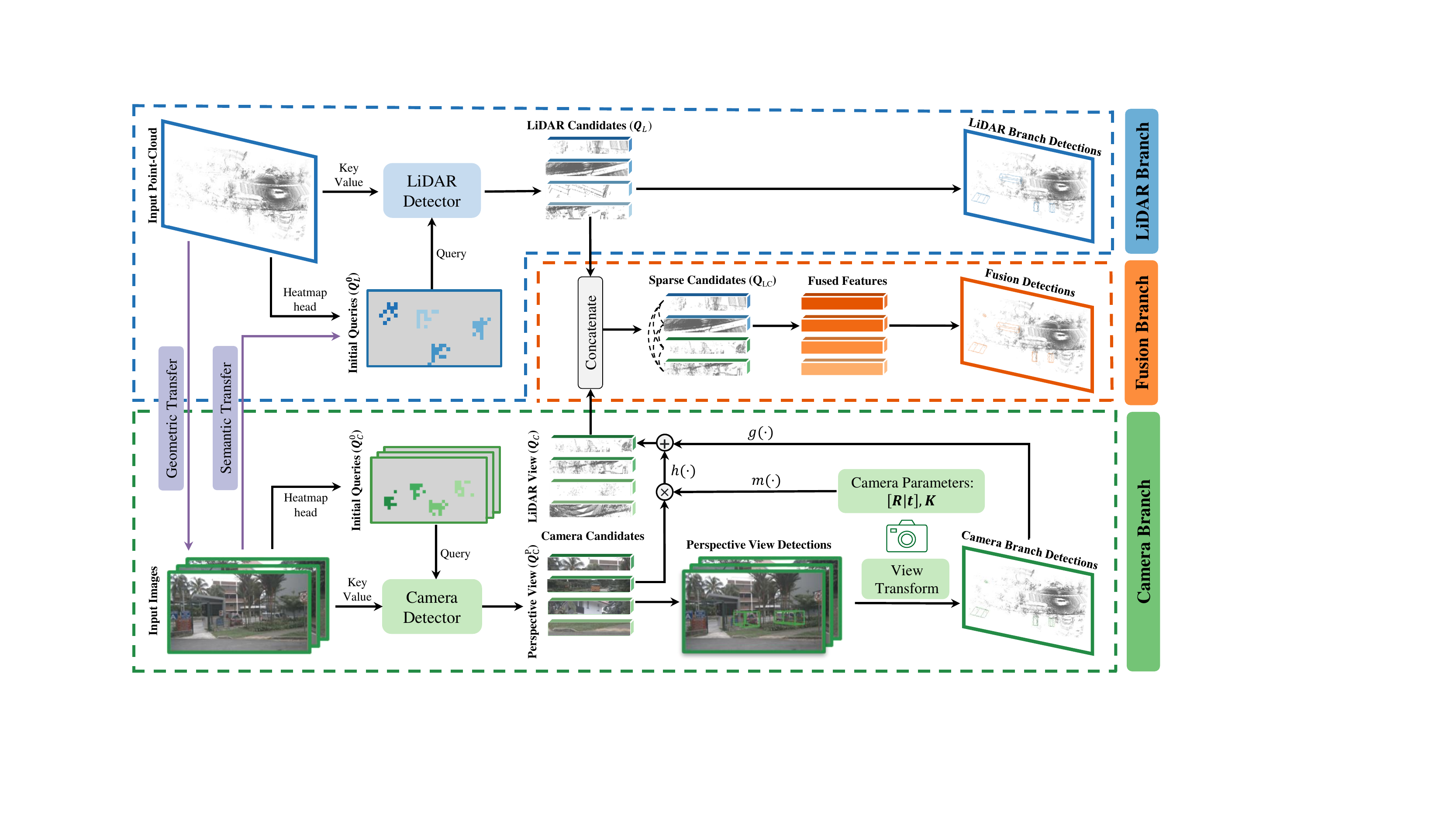}
    \vspace{-10pt}
    \caption{Overall framework of SparseFusion. It fuses sparse candidates from LiDAR and camera modalities to obtain a multi-modality instance-level representation in the unified LiDAR space.}
    \label{fig:arch}
    \vspace{-10pt}
\end{figure*}

\section{Methodology}
We present SparseFusion, an effective and efficient framework for 3D object detection via LiDAR and camera inputs. The overall architecture is illustrated in Fig.~\ref{fig:arch}. We acquire sparse candidates from each modality using modality-specific object detection in the LiDAR and camera branches. The instance-level features generated by the camera branch are transformed into the LiDAR space of the instance-level features generated by the LiDAR branch. They are then fused with a simple self-attention module (Sec.~\ref{sec:fuse}). To mitigate the negative transfer between modalities, we apply a geometric transfer module and a semantic transfer module prior to the parallel detection branches (Sec.~\ref{sec:transfer}). Furthermore, we design custom loss functions for each module to ensure stable optimization (Sec.~\ref{sec:loss}). 

% The scope would be: 3.1: Sparse Representations fusion --- 2D and 3D detectors, query initializaiton, camera view transformation, fusion by self-attention.  3.2. Cross-modality information transfer. 3.3 Loss functions for model optimization.

% the geometric and semantic information is transferred across LiDAR and camera modalities (Sec.~\ref{sec:transfer}) in the input stage of modality-specific detectors. We design loss functions for each module to ensure stable optimization (Sec.~\ref{sec:loss}). 

\subsection{Sparse Representation Fusion}
\label{sec:parallel}
% For LiDAR modality, objects are directly detected in the LiDAR coordinate. For camera modality, we first recognize objects in the perspective view of each camera and transform them into the LiDAR coordinate in line with the LiDAR modality. 

\noindent\textbf{Acquiring candidates in two modalities.}
\label{sec:lidar_detection}
\textit{LiDAR candidates.} We follow TransFusion-L~\cite{bai2022transfusion} and use only one decoder layer for LiDAR modality detection. The LiDAR backbone extracts a BEV feature map from the point cloud inputs. We initialize $N_L$ object queries $\mathbf{Q}_L^{0}=\{\mathbf{q}^{0}_{L,i}\}_{i=1}^{N_L}, \mathbf{q}^{0}_{L,i}\in\mathbb{R}^C$ as well as their corresponding reference points $\mathbf{p}_L^{0}=\{\mathbf{p}^{0}_{L,i}\}_{i=1}^{N_L}, \mathbf{p}^{0}_{L,i}\in\mathbb{R}^2$ in the BEV plane. These queries interact with the BEV features through a cross-attention layer to generate the updated object queries $\mathbf{Q}_L=\{\mathbf{q}_{L,i}\}_{i=1}^{N_L},\mathbf{q}_{L,i}\in\mathbb{R}^C$. These updated queries $\mathbf{Q}_L$ represent the instance-level features of objects in the LiDAR modality, and we use them as the LiDAR candidates in the subsequent multi-modal fusion module. Furthermore, we apply a prediction head to each query to classify the object and regress the bounding box in LiDAR coordinate space.
% , including its location in the BEV plane, height, orientation, scale, and velocity. 

\noindent\textit{Camera candidates.}
\label{sec:camera_detection}
% Despite recent efforts to directly detect objects in the 3D view by combining multi-camera inputs~\cite{wang2022detr3d,liu2022petr}, this line of work typically requires many (\textit{e.g., 6}) stacked decoder layers to do the implicit view transformation between multiple cameras. In pursuit of higher efficiency, we still resort to the monocular object detector. In this case, objects are detected in the perspective view of each camera, and multi-camera results are combined explicitly in Sec.~\ref{sec:transform}.
To generate the camera candidates, we utilize a camera-only 3D detector with images from different perspective views as inputs. Specifically, we extend deformable-DETR~\cite{zhu2020deformable} with 3D box regression heads.
% The multi-scale image features borrow depth information from LiDAR in Sec.~\ref{sec:geometric} to get the depth-aware image features. 
We also initialize $N_C$ object queries $\mathbf{Q}_C^{0}=\{\mathbf{q}^{0}_{C,i}\}_{i=1}^{N_C},\mathbf{q}^{0}_{C,i}\in\mathbb{R}^C$ along with their corresponding reference points $\mathbf{p}_C^{0}=\{\mathbf{p}^{0}_{C,i}\}_{i=1}^{N_C},\mathbf{p}^{0}_{C,i}\in\mathbb{R}^2$ on the image. For each perspective view $v$, queries on its image interact with the corresponding image features using a deformable attention layer~\cite{zhu2020deformable}. The outputs of all perspective views comprise 
the updated queries $\mathbf{Q}_C^{P}=\{\mathbf{q}^{P}_{C,i}\}_{i=1}^{N_C},\mathbf{q}^{P}_{C,i}\in\mathbb{R}^C$. We use these queries as the camera candidates in the subsequent multi-modal fusion module. We provide further details of our architecture, the initialization method, and the prediction heads for the two modalities in our supplementary materials.

\noindent\textbf{Transformation}
\label{sec:transform}
After acquiring the candidates from each modality, we aim to transform the candidates from the camera modality to the space of the candidates from the LiDAR modality. 
 Since the candidates from the camera modality are high-dimensional latent features that are distributed differently than the candidates from the LiDAR modality, a naive coordinate transformation between modalities is inapplicable here. To address this issue, we disentangle the representations of the camera candidates. Intrinsically, a camera candidate is an instance feature that is a representation of a specific object's class and 3D bounding box. While an object's class is view-invariant, its 3D bounding box is view-dependent. This motivates us to focus on transforming high-dimensional bounding box representations. 

We first input the candidate instance features into the prediction head of the camera branch. We label the outputted bounding boxes as $\mathbf{b}^P$.  Given the extrinsic matrix $[\mathbf{R}_v|\mathbf{t}_v]$ and intrinsic matrix $\mathbf{K}_v$ of the corresponding $v$-th camera, the bounding boxes can be easily projected into the LiDAR coordinate system. We denote the project bounding boxes as $\mathbf{b}^L$. We encode the projected bounding boxes with a multi-layer perceptron (MLP) $g(\cdot)$, yielding a high-dimensional box embedding. We also encode the flattened camera parameter with another MLP $m(\cdot)$ to obtain a camera embedding. The camera embedding is multiplied with the original instance features as done in \cite{li2022bevdepth}, which are then added to the box embedding, given by \begin{equation}
    \mathbf{q}_{C,i}^L=g(\mathbf{b}^L_i) + h(\mathbf{q}_{C,i}^P\cdot m(\mathbf{R}_v,\mathbf{t}_v,\mathbf{K}_v)),
    \label{eq:view}
\end{equation}
where $h(\cdot)$ is an extra MLP to encode the query features in the perspective view. $h(\cdot)$ aims to preserve view-agnostic information while discarding view-specific information. Afterward, $\mathbf{Q}_C^{L}=\{\mathbf{q}_{C,i}^L\}$ is passed through a self-attention layer to aggregate information from multiple cameras to get the updated queries $\mathbf{Q}_C$ which represent the image modality instance features in the LiDAR space.

\noindent\textbf{Sparse candidate fusion.}
\label{sec:fuse}
Our parallel modality-specific object detection provides sparse instance candidates $\mathbf{Q}_L=\{\mathbf{q}_{L,i}\}_{i=1}^{N_L}$ and $\mathbf{Q}_C=\{\mathbf{q}_{C,i}\}_{i=1}^{N_C}$ from the LiDAR and camera modalities respectively. After the above transformation of the camera candidates into LiDAR space, candidates from both modalities represent bounding boxes in the same LiDAR coordinate space, as well as the view-invariant categories. We now concatenate the candidates together:
% \begin{equation}
% \begin{split}
%     \mathbf{Q}_{LC}&=\{\mathbf{q}_{LC,i}\}_{i=1}^{N_L+N_C}\\
%     &=\{f_L(\mathbf{q}_{L,i})\}_{i=1}^{N_L}\cup \{f_C(\mathbf{q}_{C,i})\}_{i=1}^{N_C},
% \end{split}
% \end{equation}
\begin{equation}
\resizebox{0.9\linewidth}{!}{
    $\mathbf{Q}_{LC}=\{\mathbf{q}_{LC,i}\}_{i=1}^{N_L+N_C}=\{f_L(\mathbf{q}_{L,i})\}_{i=1}^{N_L}\cup \{f_C(\mathbf{q}_{C,i})\}_{i=1}^{N_C}
$}
\end{equation}
where $f_L(\cdot),f_C(\cdot)$ are learnable projectors. Afterward, we make novel use of a self-attention module to fuse the two modalities. Despite the simplicity of self-attention, the inherent intuition is novel: the modality-specific detectors encode the advantageous aspects of their respective inputs, and the self-attention module is able to aggregate and preserve the information from both modalities in an efficient manner. The output of the self-attention module is used for final classification and regression of the bounding boxes.

% $\mathbf{Q}_{LC}$ is fed into a self-attention module and feed-forward network to aggregate multi-modality instance-level information. The output $\mathbf{Q}_{F}$ is used to classify the object and regress the bounding box in the LiDAR coordinate with a prediction head same as LiDAR modality detection (Sec.~\ref{sec:lidar_detection}).

\begin{figure}[t!]
    \centering
    \includegraphics[width=1.0\linewidth]{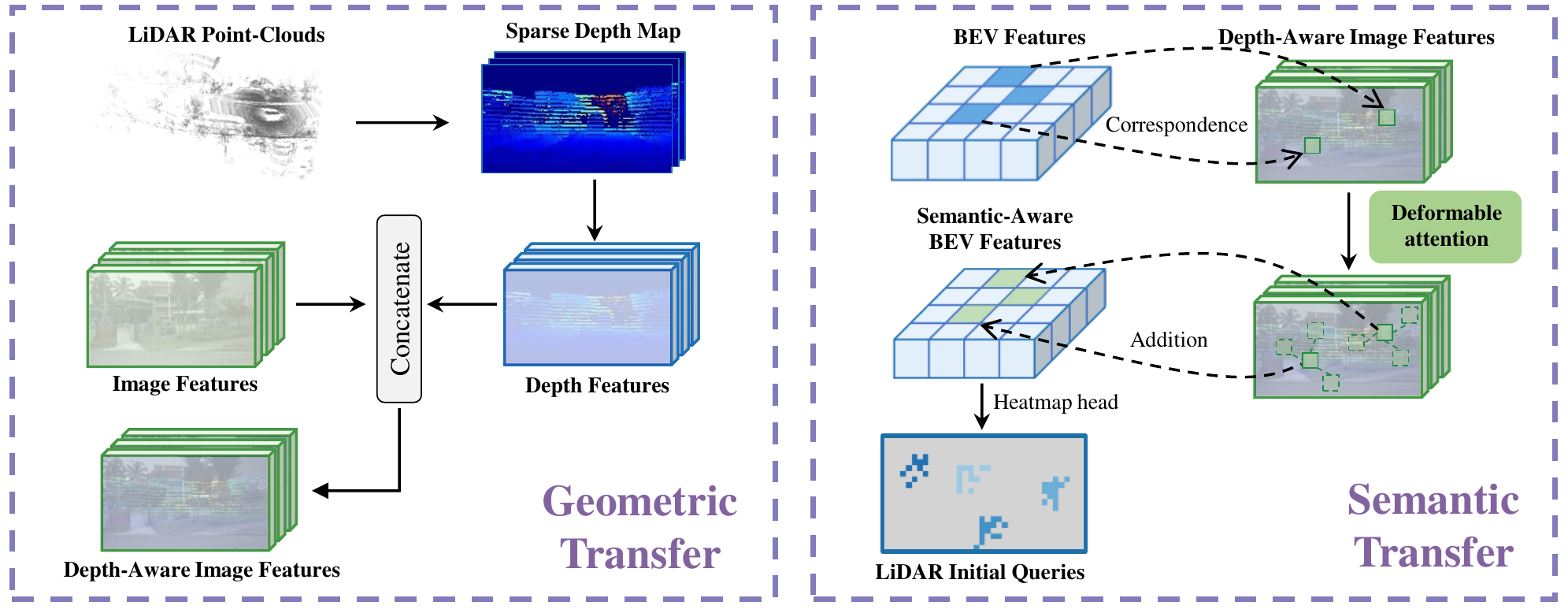}
    \caption{Cross-modality information transfer. We transfer geometric information from LiDAR to camera modality, and semantic information from camera to LiDAR modality.}
    \label{fig:transfer}
    \vspace{-10pt}
\end{figure}

\subsection{Cross-Modality Information Transfer}
\label{sec:transfer}
Although we aim to utilize the advantages of both modalities, we must address that the modalities also have their own disadvantages that can result in negative transfer between modalities. For example, the LiDAR detector struggles to capture rich semantic information, while the camera detector struggles to capture accurate geometric and depth information. To mitigate negative transfer, we propose novel geometric and semantic information transfer modules, as illustrated in Fig.~\ref{fig:transfer}, that we apply prior to the modality-specific detectors.

% A prerequisite for sparse instance-level fusion is to make the independent LiDAR and camera modality-specific detection accurate and robust enough. However, LiDAR inputs require complementary semantics, and the camera modality lacks 3D geometric information. To this end, we transfer these information across modalities. 

\noindent\textbf{Geometric transfer from LiDAR to camera.}
\label{sec:geometric}
We project each point in the LiDAR point cloud input to multi-view images to generate sparse multi-view depth maps. These multi-view depth maps are inputted into a shared encoder to obtain depth features, which are then concatenated with the image features to form depth-aware image features that compensate for the lack of geometric information in camera inputs. The depth-aware image features are used as the input to the camera branch.

% We project each point $(x,y,z)$ in LiDAR point cloud to multi-view images to generate the sparse depth maps $\mathbf{D}_C=\{\mathbf{D}_{C,v}\}$, where $v=1,2,\dots, V$ refer to different views. These multi-view depth maps are fed into a shared encoder $q(\cdot)$ to obtain the multi-scale depth feature maps $\{\mathbf{F}_{D,v}^l\}_{l=1}^L=q(\mathbf{D}_{C,v})$. The depth features are concatenated with the image features $\mathbf{F}_C=\{\mathbf{F}_{C,v}^l\}$ as the depth-aware image features $\mathbf{\hat{F}}_C=\{\mathbf{\hat{F}}_{C,v}^l\}$, where $\mathbf{\hat{F}}_{C,v}^l=Concat(\mathbf{F}_{C,v}^l,\mathbf{F}_{D,v}^l)$. The depth-aware image features serve as the input of the camera detector in Sec.~\ref{sec:camera_detection} to make up for the geometric weakness of camera inputs.

\noindent\textbf{Semantic transfer from camera to LiDAR.}
\label{sec:semantic}
We project the points in the LiDAR point cloud input to the image inputs, which yields sparse points on the image features. We perform max-pooling to aggregate the resulting multi-scale features, and we combine them with the BEV feature through addition. The concatenated features serve as the queries and interact with the multi-scale image features through deformable-attention~\cite{zhu2020deformable}. The updated queries replace the original queries in the BEV features, which results in the semantic-aware BEV features, which are used for query initialization.
% while the original BEV features are used as the input to the off-the-shelf LiDAR detector to maximize performance.

\subsection{Objective Function}
\label{sec:loss}
We apply the Gaussian focal loss~\cite{lin2017focal} to the initialized queries of both modalities, given by
\begin{equation}
    \mathcal{L}_{init}=\mathcal{L}_{GFocal}(\mathbf{\hat{Y}}_{L},\mathbf{Y}_L)+\mathcal{L}_{GFocal}(\mathbf{\hat{Y}}_{C},\mathbf{Y}_C),
\end{equation}
where $\mathbf{\hat{Y}}_{L},\mathbf{\hat{Y}}_{C}$ are the dense predictions of category-wise heatmaps of the LiDAR and camera modalities, respectively, and $\mathbf{Y}_{L},\mathbf{Y}_{C}$ are the corresponding ground-truths.

Then, we apply the loss function for the detectors of the LiDAR and camera modalities, as well as the view transformation of the camera candidates and the candidate fusion stage. Firstly, the predictions of each modality-specific detector are independently matched with the ground-truth using the Hungary algorithm~\cite{kuhn1955hungarian}. The object classification is optimized with focal loss~\cite{lin2017focal} and the 3D bounding box regression is optimized with L1 loss. For the camera modality detector, the ground-truth bounding boxes are in separate camera coordinates. For all other detectors, ground-truth bounding boxes are in LiDAR coordinates. The detection loss can be represented as
\begin{equation}
\mathcal{L}_{detect}=\gamma\mathcal{L}_{detect}^{camera}+\mathcal{L}_{detect}^{trans}+\mathcal{L}_{detect}^{LiDAR}+\mathcal{L}_{detect}^{fusion}.
\end{equation}
Our entire network is optimized using $\mathcal{L}=\alpha\mathcal{L}_{init}+\beta\mathcal{L}_{detect}$. In our implementation, we empirically set $\gamma=1, \alpha=0.1$, and $\beta=1$ to balance different terms.

\begin{table*}[htb!]
    \centering
    \caption{Comparison with existing methods on nuScenes \textit{validation set} and \textit{test set}.}
    \label{tab:comparison}
    \vspace{-10pt}
    \begin{tabular}{c|c|cc|cc|cc}
    \toprule
        \multirow{2}{*}{Methods} & \multirow{2}{*}{Modality} & \multirow{2}{*}{LiDAR Backbone} & \multirow{2}{*}{Camera Backbone} & \multicolumn{2}{c|}{\textit{validation set}} & \multicolumn{2}{c}{\textit{test set}}\\
        & & & & NDS & mAP & NDS & mAP\\
        \midrule
        FCOS3D~\cite{wang2021fcos3d} & Camera & - & ResNet-101~\cite{he2016deep} & 41.5 & 34.3 &  42.8 & 35.8\\
        PETR~\cite{liu2022petr} & Camera & - & ResNet-101~\cite{he2016deep} & 44.2 & 37.0 & 45.5 & 39.1\\
        \midrule
        CenterPoint~\cite{yin2021center} & LiDAR & VoxelNet~\cite{zhou2018voxelnet} & - &  66.8 & 59.6 & 67.3 & 60.3\\
        TransFusion-L~\cite{bai2022transfusion}& LiDAR & VoxelNet~\cite{zhou2018voxelnet} & - & 70.1 & 65.1 & 70.2 & 65.5\\ 
        \midrule
        PointAugmenting~\cite{vora2020pointpainting} & LiDAR+Camera & VoxelNet~\cite{zhou2018voxelnet} & DLA34~\cite{yu2018deep} & - & - & 71.0 & 66.8\\

        FUTR3D~\cite{chen2022futr3d} & LiDAR+Camera & VoxelNet~\cite{zhou2018voxelnet} & ResNet-101~\cite{he2016deep} & 68.3 & 64.5 & - & -\\
        UVTR~\cite{li2022unifying} & LiDAR+Camera & VoxelNet~\cite{zhou2018voxelnet} & ResNet-101~\cite{he2016deep} & 70.2 & 65.4 & 71.1 & 67.1\\
        TransFusion~\cite{bai2022transfusion} & LiDAR+Camera & VoxelNet~\cite{zhou2018voxelnet} & ResNet-50~\cite{he2016deep} & 71.3 & 67.5 & 71.6 & 68.9\\
        AutoAlignV2~\cite{chen2022autoalignv2} & LiDAR+Camera & VoxelNet~\cite{zhou2018voxelnet} & CSPNet~\cite{wang2020cspnet} & 71.2 & 67.1 & 72.4 & 68.4\\
        % BEVFusion~\cite{liang2022bevfusion} & LiDAR+Camera & VoxelNet~\cite{zhou2018voxelnet} & Dual-Swin-T & 71.0 & 67.9 & 71.8 & 69.2\\
        BEVFusion~\cite{liang2022bevfusion} & LiDAR+Camera & VoxelNet~\cite{zhou2018voxelnet} & Dual-Swin-T~\cite{liang2022cbnet} & 72.1 & 69.6 & 73.3 & 71.3\\
        BEVFusion~\cite{liu2022bevfusion} & LiDAR+Camera & VoxelNet~\cite{zhou2018voxelnet} & Swin-T~\cite{liu2021swin} & 71.4 & 68.5 & 72.9 & 70.2\\
        DeepInteraction~\cite{yang2022deepinteraction} & LiDAR+Camera & VoxelNet~\cite{zhou2018voxelnet} & ResNet-50~\cite{he2016deep} & 72.6 & 69.9 & 73.4 & 70.8\\
        % CMT~\cite{yan2023cross} & LiDAR+Camera & VoxelNet~\cite{zhou2018voxelnet} & VoV-99~\cite{lee2020centermask} & 71.9 & 69.4 & 73.0 & 70.4\\
        CMT~\cite{yan2023cross} & LiDAR+Camera & VoxelNet~\cite{zhou2018voxelnet} & ResNet-50~\cite{he2016deep} & 70.8 & 67.9 & - & -\\
        \midrule
        \textbf{SparseFusion (ours)} & LiDAR+Camera & VoxelNet~\cite{zhou2018voxelnet} & ResNet-50~\cite{he2016deep} & \textbf{72.8} & \textbf{70.4} & \textbf{73.8} & \textbf{72.0}\\
        \bottomrule
    \end{tabular}
\end{table*}

\begin{figure*}[t!]
    \centering
    \includegraphics[width=0.9\linewidth]{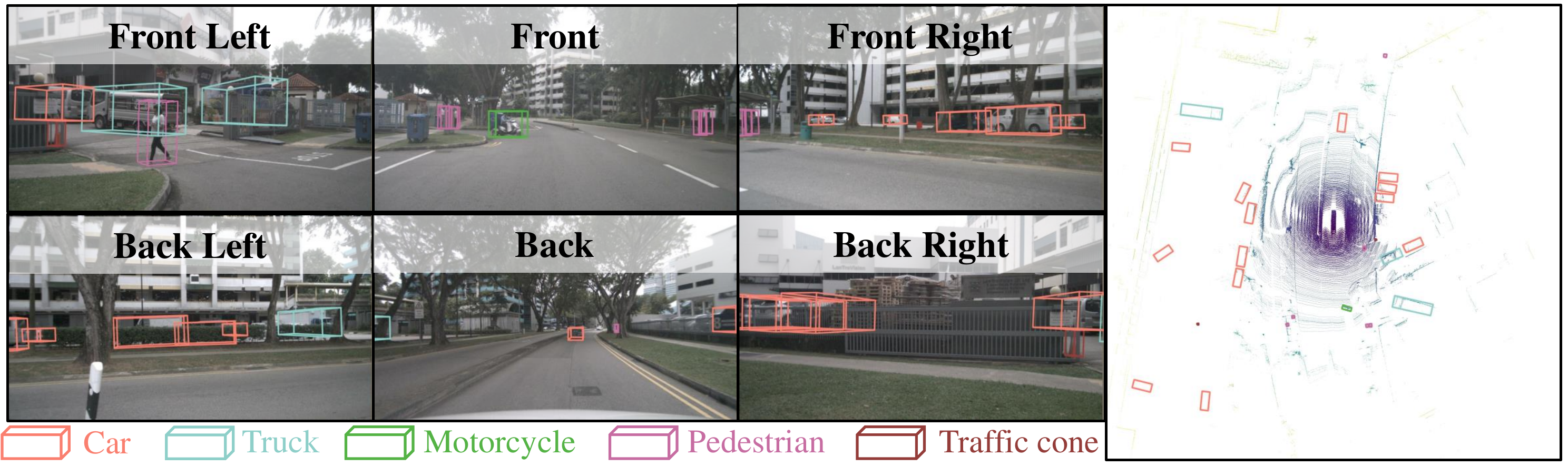}
    \caption{Qualitative results of SparseFusion on nuScenes \textit{validation set}.}
    \label{fig:visualize}
    \vspace{-10pt}
\end{figure*}

\section{Experiments}
\subsection{Dataset and metrics}
We follow previous work~\cite{yang2022deepinteraction,yan2023cross,liang2022bevfusion,chen2022autoalignv2} to evaluate our method on the nuScenes dataset~\cite{caesar2020nuscenes}. It is a challenging dataset for 3D object detection, consisting of 700/150/150 scenes for training/validation/test. It provides point clouds collected using a 32-beam LiDAR and six images from multi-view cameras. There are 1.4 million annotated 3D bounding boxes for objects from 10 different classes. We evaluate performance using the nuScenes detection score (NDS) and mean average precision (mAP) metrics. The final mAP is averaged over distance thresholds 0.5m, 1m, 2m, and 4m on the BEV across 10 classes. NDS is a weighted average of mAP and other true positive metrics including mATE, mASE, mAOE, mAVE, and mAAE.

\subsection{Implementation Details}
Our implementation is based on the MMDetection3D framework~\cite{mmdet3d2020}. For the camera branch, we use ResNet-50~\cite{he2016deep} as the backbone and initialize it with the Mask R-CNN~\cite{he2017mask} instance segmentation network pretrained on nuImage~\cite{caesar2020nuscenes}. The input image resolution is $800$$\times$$448$. For the LiDAR branch, we apply VoxelNet~\cite{zhou2018voxelnet} with voxel size $(0.075m, 0.075m, 0.2m)$. The detection range is set as $[-54m, 54m]$ for the XY-axes and $[-5m,3m]$ for the Z-axis. Our LiDAR and camera modality detectors both include only 1 decoder layer. The query numbers for the LiDAR and camera modalities are set as $N_L=N_C=200$, so our fusion stage can detect at most $400$ objects per scene.

Since our framework disentangles the camera detector and the LiDAR detector, we can conveniently apply data augmentation separately to the LiDAR inputs and camera inputs. We apply random rotation, scaling, translation, and flipping to the LiDAR inputs, and we apply random scaling and horizontal flipping to the camera inputs. Our training pipeline follows previous works~\cite{bai2022transfusion,liu2022bevfusion,yang2022deepinteraction}. We first train TransFusion-L~\cite{bai2022transfusion} as our LiDAR-only baseline, which is used to initialize our LiDAR backbone and LiDAR modality detector. This LiDAR-only baseline is trained for 20 epochs. Afterward, we freeze the pretrained LiDAR components and train the entire fusion framework for 6 epochs. For both training stages, we use the AdamW optimizer~\cite{loshchilov2017decoupled} with one-cycle learning rate policy~\cite{smith2017cyclical}. The initial learning rate is $10^{-4}$ and the weight decay is $10^{-2}$. The hidden dimensions in the entire model except the backbones are $128$. For both training stages, we adopt CBGS~\cite{zhu2019class} to balance the class distribution. We train our method on four NVIDIA A6000 GPUs with batch size 16.
% \begin{figure}[t]
%     \centering
%     \begin{subfigure}{0.49\linewidth}
%          \centering
%          \includegraphics[width=\linewidth]{figures/FPS_mAP.pdf}
%          \caption{mAP \textit{v.s.} FPS}
%      \end{subfigure}
%      \hfill
%      \begin{subfigure}{0.49\linewidth}
%          \centering
%          \includegraphics[width=\linewidth]{figures/FPS_NDS.pdf}
%          \caption{NDS \textit{v.s.} FPS}
%      \end{subfigure}
%     \caption{Trade-off between efficiency and performance (\textit{test set}). $\dag:$ Official code of \cite{chen2022autoalignv2} uses flip as test-time augmentation. $\ddag:$ We use BEVFusion-base results in the official repository of \cite{liu2022bevfusion} to match input resolutions of other methods. $\S:$ Swin-T~\cite{liu2021swin,liang2022cbnet} is adopted as image backbone.}
%     \label{fig:trade-off}
% \end{figure}

\subsection{Results and Comparison}
% \noindent\textbf{Comparison with Existing Methods.}
We report results on the nuScenes \textit{validation} and \textit{test sets}, without using any test-time augmentations or model ensembles. As shown in Tab.~\ref{tab:comparison}, SparseFusion significantly improves our LiDAR-only baseline, TransFusion-L~\cite{bai2022transfusion}, by +3.6\% NDS and +6.3\% mAP on the \textit{test set}, due to the additional use of camera inputs. More importantly, SparseFusion sets a new state of the art on both the \textit{validation set} and \textit{test set}, outperforming prior works including those using stronger backbones. It is noteworthy that SparseFusion demonstrates a 0.4\% NDS and 1.0\% mAP improvement over the most recent state-of-the-art~\cite{yang2022deepinteraction} while also achieving a 1.8x speedup (5.6 FPS vs. 3.1 FPS) on A6000 GPU. It can be seen in Fig.~\ref{fig:trade-off} that, in addition to superior performance, SparseFusion also provides the fastest inference speed. We also demonstrate the performance of SparseFusion by visualizing some qualitative results in Fig.~\ref{fig:visualize}.

% We believe that the improvement and acceleration attribute to the sparse representations, which enables more efficient interaction between modalities and casts the advantages of both modalities. To shed light on why it works, we conduct extensive analysis and ablation studies below.

% \noindent\textbf{Efficiency and Model Complexity}
% Our model-specific detectors represent instance features in the unified LiDAR coordinate to reduce the domain gaps. This allows for an efficient attention-based sparse fusion strategy, which only costs negligible time. As a result, our method achieves a great trade-off between efficiency and effectiveness. As shown in Fig.~\ref{fig:trade-off}, apart from the best performance on the nuScenes \textit{test set}, SparseFusion also has great superiority in inference speed.

\subsection{Analysis}
\noindent\textbf{Performance Breakdown}
SparseFusion performs 3D object detection in both the LiDAR and camera branch separately. Tab.~\ref{tab:breakdown} shows the detection performance in different parts of SparseFusion including the LiDAR branch, camera branch (before and after view transformation), and the fusion branch. We notice that our LiDAR branch detection results notably surpass the LiDAR-only baseline TransFusion-L~\cite{bai2022transfusion} since the proposed semantic transfer can compensate for the weakness of point cloud inputs. In comparison with the state-of-the-art single-frame camera detector PETR~\cite{liu2022petr} (six decoder layers), our camera branch achieves much better performance with just one decoder layer, owing to the depth-aware features from the proposed geometric transfer. Besides, our view transformation module not only transforms the instance features from the camera coordinate space into the LiDAR coordinate space, but it also slightly improves the camera branch detection performance by aggregating multi-view information. With this strong performance, the modality-specific detectors in each branch would not cause negative transfer during fusion.

\begin{table}[htb!]
    \centering
    \caption{Performance breakdown. We show the detection results in different parts of SparseFusion on the \textit{validation set}. 'L' and 'C' refer to the LiDAR and camera modalities, respectively. 'ST' and 'GT' refer to semantic transfer and geometric transfer, respectively. 'VT' is the view transformation module.}
    \label{tab:breakdown}
    \begin{tabular}{cccc}
        \toprule
        Methods & Modality & NDS & mAP\\
        \midrule
        TransFusion-L~\cite{bai2022transfusion} & L & 70.1 & 65.1\\
        LiDAR branch & L + ST & \textbf{71.8} & \textbf{68.4}\\
        \midrule
        PETR (ResNet50)~\cite{liu2022petr} & C & 38.1 & 31.3\\
        Camera branch (before VT) & C+GT & 43.5 & 40.6 \\
        Camera branch (after VT) & C+GT & \textbf{44.3} & \textbf{41.5}\\
        \midrule
        SparseFusion & L+C & \textbf{72.8} & \textbf{70.4}\\
        \bottomrule
    \end{tabular}
\end{table}

\noindent\textbf{Strong Image Backbone.} We incorporate stronger Swin-T~\cite{liu2021swin} backbone into SparseFusion to match some previous work~\cite{liang2022bevfusion,yan2023cross}. Tab.~\ref{tab:backbone} compares the performance of different methods on the nuScenes \textit{validation set}. We do not include any test-time augmentations or model ensembles. Although multi-modality detection relies more on the LiDAR inputs, SparseFusion can still benefit from a stronger image backbone and beat all the counterparts.
\begin{table}[htb!]
    \centering
    \caption{Results on nuScenes \textit{validation set} with stronger image backbones. We do not use any test-time augmentations or model ensembles. The inference speed is measured on a single NVIDIA A6000 GPU. $\dag:$ CMT~\cite{yan2023cross} adopts flash-attention~\cite{dao2022flashattention} for transformer acceleration.}
    \label{tab:backbone}
    \resizebox{\linewidth}{!}{

    \begin{tabular}{cccccc}
        \toprule
        Methods & \makecell{Image\\Backbone} & \makecell{Input\\Resolution} & NDS & mAP & FPS\\
        \midrule
        BEVFusion~\cite{liang2022bevfusion} & Dual-Swin-T & 1600$\times$900 & 72.1 & 69.6 & 0.8\\
        % CMT~\cite{yan2023cross} & ResNet-50~\cite{he2016deep} & 800$\times$320 & 70.8 & 67.9\\
        CMT~\cite{yan2023cross} & VoVNet-99 & 1600$\times$640 & \underline{72.9} & 70.3 & 3.8$^{\dag}$\\
        \midrule
        SparseFusion & ResNet-50 & 800$\times$448 & 72.8 & \underline{70.4} & \textbf{5.6}\\
        SparseFusion & Swin-T & 800$\times$448 & \textbf{73.1} & \textbf{71.0} & \underline{5.3}\\
        \bottomrule
    \end{tabular}
    }

\end{table}

\begin{figure}[t!]
    \centering
    \begin{subfigure}{0.48\linewidth}
         \centering
         \includegraphics[width=\linewidth]{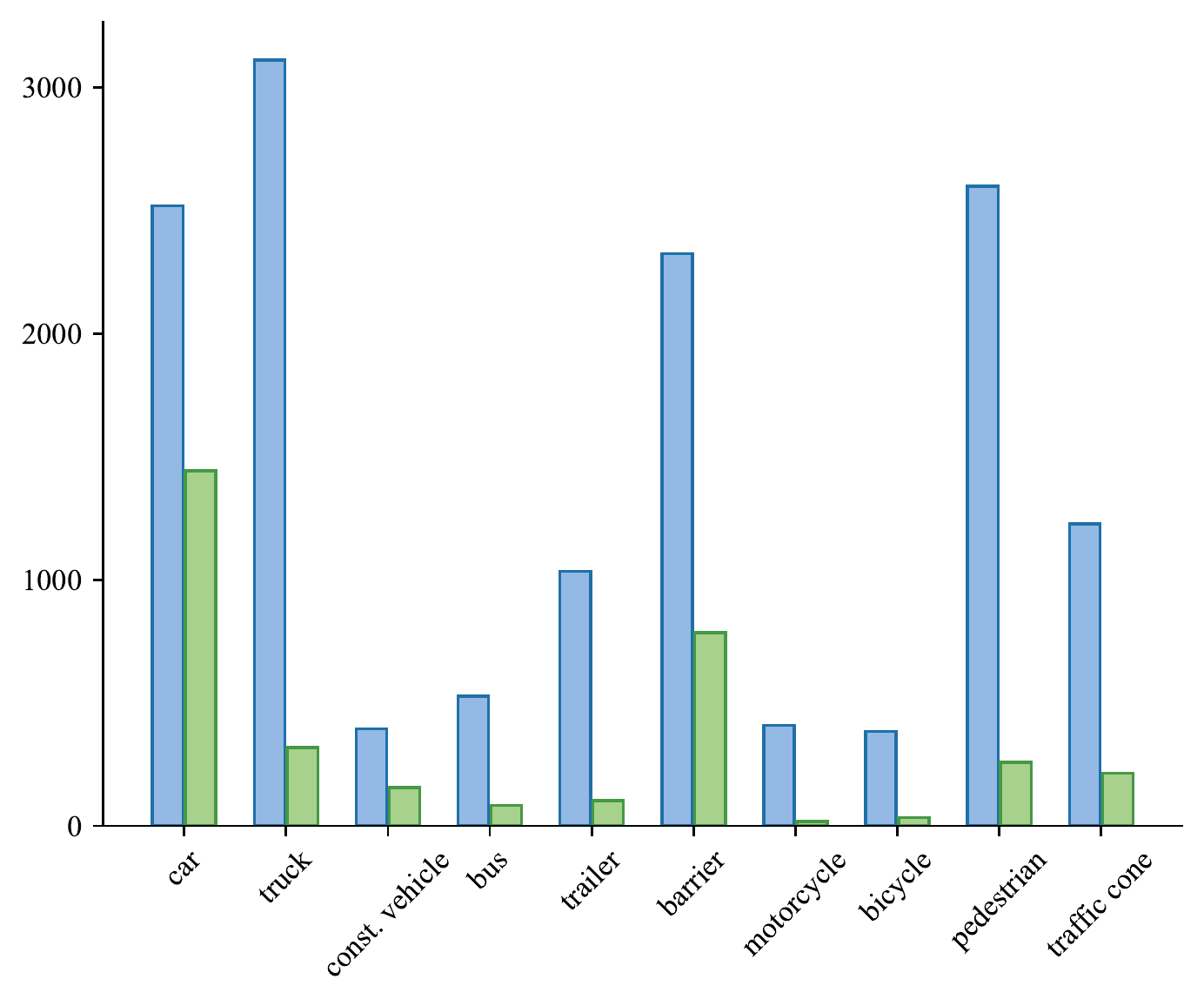}
         \caption{Objects of different classes.}
         \label{fig:recall-category}
     \end{subfigure}
     \hfill
     \begin{subfigure}{0.5\linewidth}
         \centering
         \includegraphics[width=\linewidth]{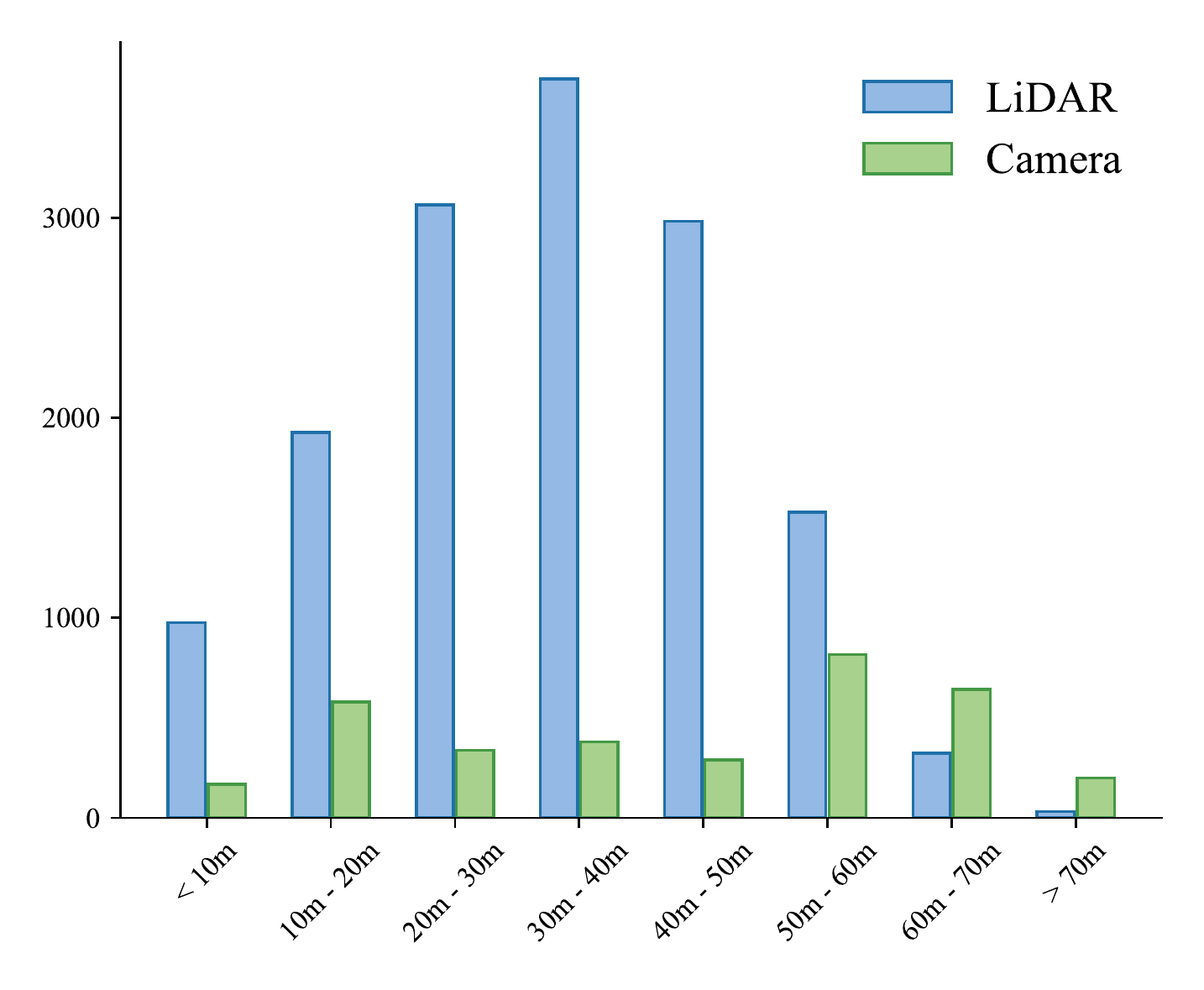}
         \caption{Objects at different distances.}
         \label{fig:recall-distance}
     \end{subfigure}
     \vspace{-5pt}
    \caption{Number of objects recalled by each modality. We do not count those objects recalled by both or neither.}
    \label{fig:recall}
    \vspace{-10pt}
\end{figure}

\noindent\textbf{Modality-Specific Object Recall.} The parallel detectors in the LiDAR and camera branches enable us to determine which modality recalls each object. Given $N_L+N_C$ predictions, we know that the first $N_L$ instances come from the LiDAR-modality detector, while the last $N_C$ are from the camera modality. An object is recalled if a bounding box with correct classification is predicted within \textit{a radius of two meters} around it. In Fig.~\ref{fig:recall}, we demonstrate the number of objects in the nuScenes \textit{validation set} recalled by exactly one modality. We observe that each modality can compensate for the weakness of the other to some extent. Although the LiDAR modality is typically more powerful, the camera modality plays an important role in detecting objects from classes such as cars, construction vehicles, and barriers. Furthermore, the camera modality is useful for detecting objects at far distances where point clouds are sparse.

\noindent\textbf{Cross-Modality Sparse Representation Interaction.} Fig.~\ref{fig:attention} visualizes the instance feature interaction in the sparse fusion stage. The strength of attention between instance-level features is reflected by the thickness and darkness of lines. We notice that most objects can aggregate multi-modality instance-level features during fusion. Although the strongest interactions exist mainly between neighboring instances, it is interesting that features from the camera modality are also able to deliver strong interactions with the instances at the distant range. 
% This could be a result of objects of the same class sharing very similar semantic details in images.
This could be a result of the shared semantics among objects in images.

\begin{figure}[ht!]
    \centering
    \vspace{-10pt}
    \includegraphics[width=0.48\linewidth]{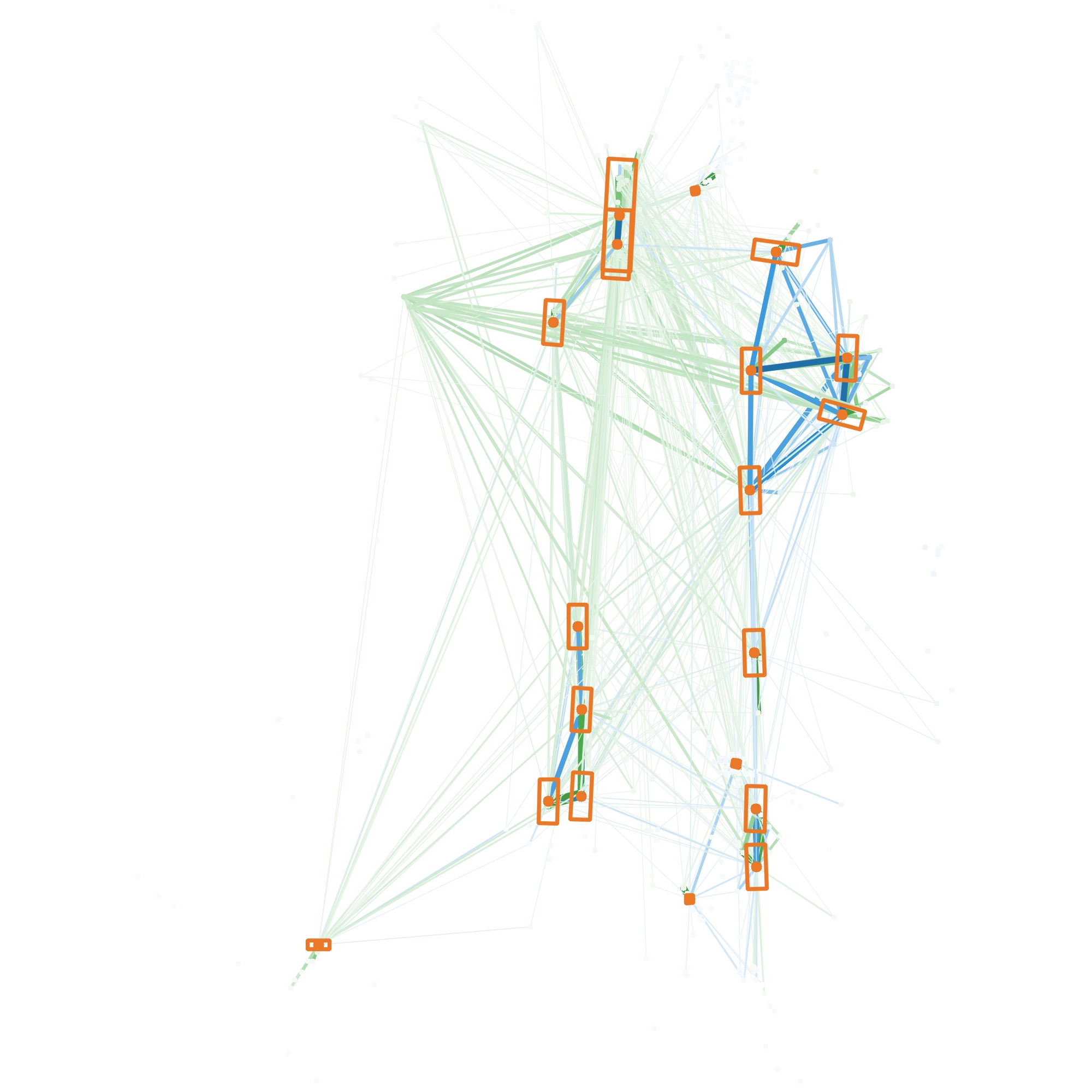}
    \hfill
    \includegraphics[width=0.48\linewidth]{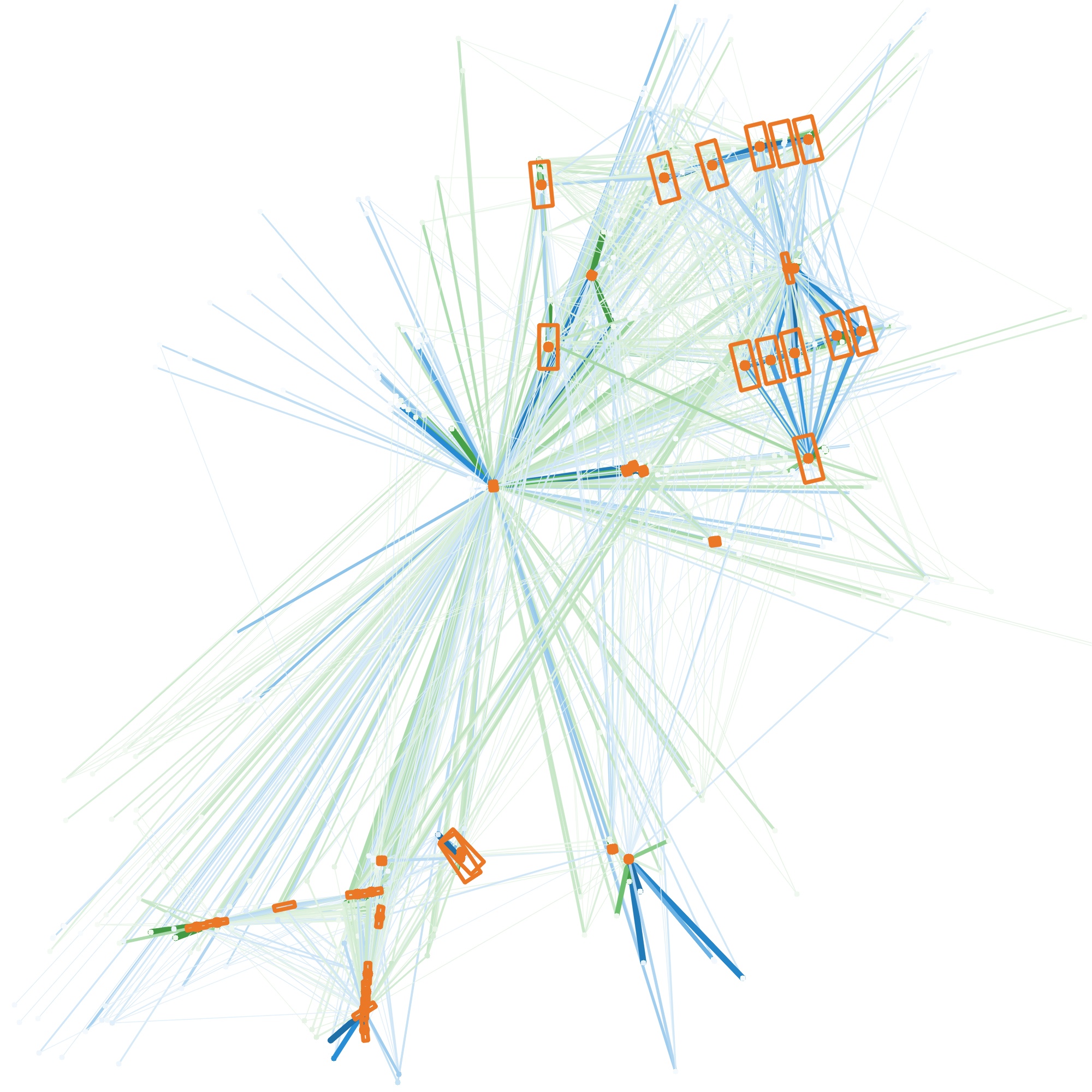}
    \includegraphics[width=0.48\linewidth]{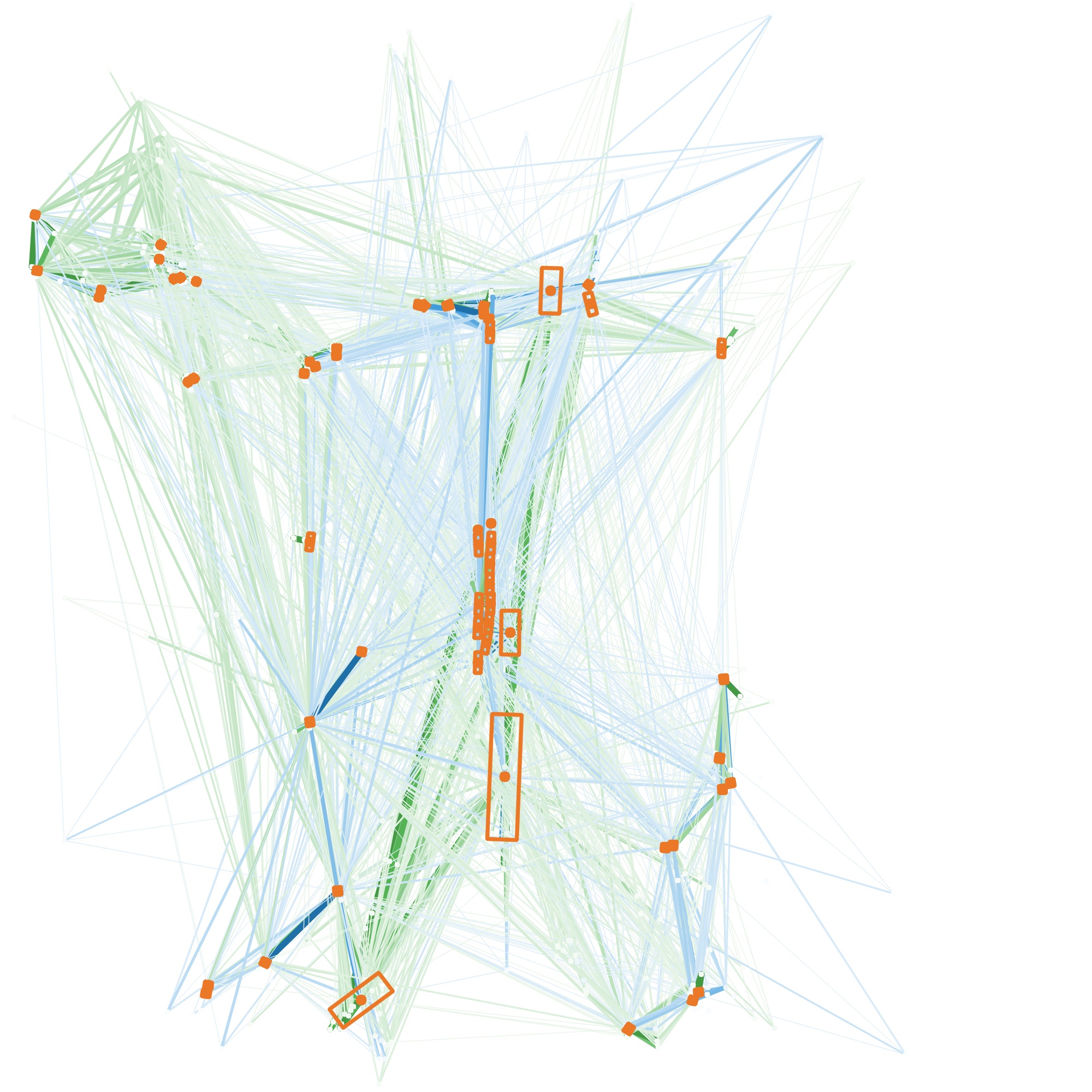}
    \hfill
    \includegraphics[width=0.48\linewidth]{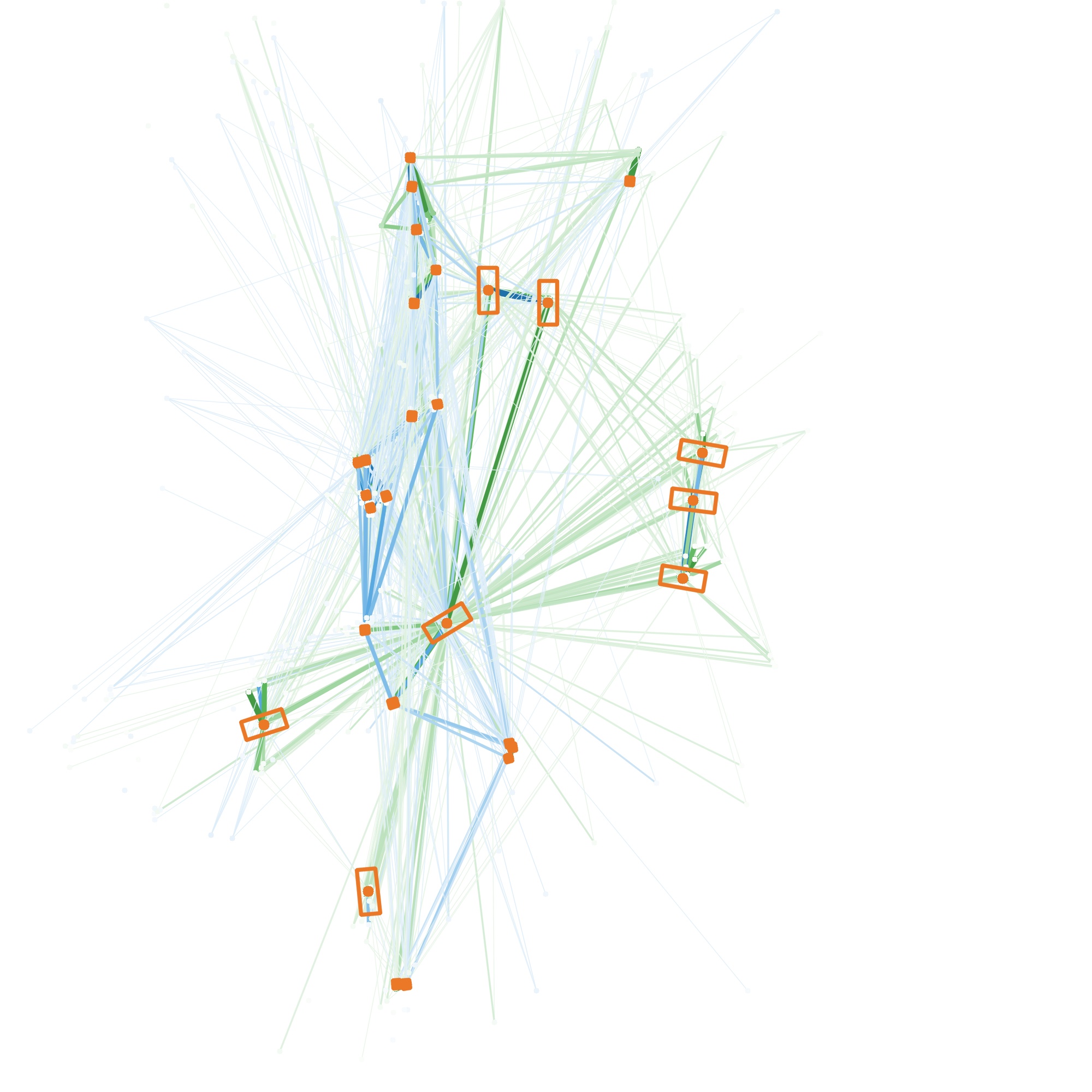}

    \vspace{-10pt}
    \caption{Instance-level feature interaction in the fusion stage. \textcolor{Orange}{Orange} boxes are detected after the fusion stage with high confidence in the BEV space. \textcolor{DarkBlue}{Blue} and \textcolor{BrightGreen}{green} dots denote all instances from the LiDAR and camera branches separately. \textcolor{Orange}{Orange} boxes are connected with \textcolor{DarkBlue}{blue}/\textcolor{BrightGreen}{green} dots with \textcolor{DarkBlue}{blue}/\textcolor{BrightGreen}{green} lines. The attention strength is represented by the darkness and thickness of the lines.} % we may add two more such figures tomorrow.
    \label{fig:attention}
    \vspace{-10pt}

\end{figure}

\begin{table*}[ht]
    \centering
    \caption{Ablation study experiments to justify our design of each module.}
    \vspace{-8pt}
    \begin{subtable}{0.33\linewidth}
        \centering
        \caption{Sparse fusion strategies.}
        \label{tab:ablate-fusion}
        \vspace{-6pt}
        \begin{tabular}{ccc}
            \toprule
            Fusion Strategy & NDS & mAP \\
            \midrule
            MLP & 67.8 & 64.5 \\
            Cross-Attention & 68.6 & 65.8 \\
            Optimal Transport & 68.0 & 65.6\\
            \midrule
            Self-Attention & \textbf{68.8} & \textbf{66.4}\\
            \bottomrule
        \end{tabular}
    \end{subtable}
    \hfill
    \begin{subtable}{0.32\linewidth}
        \centering
        \caption{Information transfers.}
        \label{tab:ablate-transfer}
        \resizebox{\linewidth}{!}{
        \begin{tabular}{cccc}
        \toprule
            Geometric & Semantic & NDS & mAP \\
        \midrule
            \CheckmarkBold & \XSolidBrush & 67.7 & 64.2 \\
            \XSolidBrush & \CheckmarkBold & 68.4 & 65.7 \\
        \midrule
         \CheckmarkBold & \CheckmarkBold & \textbf{68.8} & \textbf{66.4}\\
        \bottomrule
        \end{tabular}
        }
    \end{subtable}
    \hfill
    \begin{subtable}{0.32\linewidth}
        \centering
        \caption{Parallel modality-specific detectors.}
        \label{tab:ablate-parallel}
        \resizebox{\linewidth}{!}{
        \begin{tabular}{cccc}
            \toprule
            Seq. Pos. & Seq. Feat. & NDS & mAP \\
            \midrule
            \CheckmarkBold & \XSolidBrush & 68.1 & 65.3 \\
            \CheckmarkBold & \CheckmarkBold & 67.9 & 64.2 \\
            \midrule
            \XSolidBrush & \XSolidBrush & \textbf{68.8} & \textbf{66.4}\\
            \bottomrule
        \end{tabular}
        }
    \end{subtable}
\end{table*}

\subsection{Ablation Studies}
In this section, we study the effect of using alternatives for the different modules in SparseFusion. For our ablation studies, we train on a \textit{1/5 split} of nuScenes \textit{training set} and evaluate on the full nuScenes \textit{validation set}. 

\noindent\textbf{Sparse Fusion Strategy.} We compare our self-attention module for sparse feature fusion with other fusion methods. In addition to our self-attention module in Sec.~\ref{sec:fuse}, three alternatives are considered: 1) Instance-level candidates from the two branches are directly fed into an MLP without any cross-instance aggregation. 2) LiDAR instance candidates are used as queries and camera instance candidates are used as keys/values. A cross-attention module is then used to fuse multi-modality instance features. 3) We make novel use of optimal transport in LiDAR-camera fusion~\cite{peyre2019computational}. 
% Our hypothesis is that since candidates from two branches represent the objects in the same scene, a transformation between the candidates should be learned to map candidates in one modality to another. 
We propose to learn a distribution transformation from camera candidates to LiDAR candidates through optimal transport. Then, we can directly fuse them by concatenating the candidates of two branches along the channel dimension. More details about this method are provided in appendix. 

The results in Tab.~\ref{tab:ablate-fusion} show that only cross-attention achieves competitive performance to self-attention. Yet, we observe that it relies so heavily on the output of the LiDAR branch that the camera branch is not fully utilized to compensate for the weaknesses of the LiDAR branch. The MLP strategy has limited performance as it does not fuse cross-instance and cross-modality information. Despite the impressive progress of optimal transport in other fields, it fails to learn the correspondences between the instance features of the two modalities and thus has limited performance. In contrast, self-attention is simple, efficient, and effective.

% \begin{table}[t]
%     \centering
%     \caption{Ablation study on sparse fusion strategies.}
%     \label{tab:ablate-fusion}
%     \vspace{-10pt}
%     \begin{tabular}{ccc}
%     \toprule
%         Fusion Strategy & NDS & mAP \\
%     \midrule
%         MLP & 67.8 & 64.5 \\
%         Cross-Attention & 68.6 & 65.8 \\
%         Optimal Transport & 68.0 & 65.6\\
%     \midrule
%         Self-Attention & \textbf{68.8} & \textbf{66.4}\\
%     \bottomrule
%     \end{tabular}
% \end{table}

\noindent\textbf{Information Transfer.} We then ablate the geometric and semantic transfer between the LiDAR and camera modalities. The results in Tab.~\ref{tab:ablate-transfer} show that the fusion performance benefits from both transfers. This also validates that the disadvantages of both modalities result in negative transfer and that our proposed information transfer modules are indeed effective in mitigating this issue. The semantic transfer module especially improves the final performance since it compensates for the LiDAR modality's lack of semantic information, which is critical for 3D detection.

\noindent\textbf{View Transformation.} As explained in Sec.~\ref{sec:transform}, we transform the sparse representations of both modalities into one unified space. To validate the effectiveness of this approach, we ablate the view transformation of camera candidates into the LiDAR coordinate space. This results in a more straightforward method where we simply obtain the predictions of two modalities and directly fuse them using self-attention. Ablating the view transformation drops performance from 66.4\% mAP and 68.8\% NDS to 65.6\% mAP and 68.3\% NDS, respectively. This demonstrates that the view transformation is indeed helpful to overall performance.

% we get rid of the t metWe explicitly transform the 3D bounding boxes from the camera coordinate to the LiDAR coordinate. Without this view transformation (Sec.~\ref{sec:transform}), the performance drops to 68.3 NDS and 65.6 mAP. In this case, the fusion stage can hardly get the 3D bounding box geometry from the camera branch.

% \begin{table}[t]
%     \centering
%     \caption{Ablation study on information transfers.}
%     \label{tab:ablate-transfer}
%     \vspace{-10pt}
%     \begin{tabular}{cccc}
%     \toprule
%         Geometric & Semantic & NDS & mAP \\
%     \midrule
%         \CheckmarkBold & \XSolidBrush & 67.7 & 64.2 \\
%         \XSolidBrush & \CheckmarkBold & 68.4 & 65.7 \\
%     \midrule
%          \CheckmarkBold & \CheckmarkBold & \textbf{68.8} & \textbf{66.4}\\
%     \bottomrule
%     \end{tabular}
% \end{table}

% \begin{table}[t]
%     \centering
%     \caption{Ablation study on parallel detectors.}
%     \label{tab:ablate-parallel}
%     \vspace{-10pt}
%     \begin{tabular}{cccc}
%     \toprule
%         Seq. Pos. & Seq. Feat. & NDS & mAP \\
%     \midrule
%         \CheckmarkBold & \XSolidBrush & 68.1 & 65.3 \\
%         \CheckmarkBold & \CheckmarkBold & 67.9 & 64.2 \\
%     \midrule
%          \XSolidBrush & \XSolidBrush & \textbf{68.8} & \textbf{66.4}\\
%     \bottomrule
%     \end{tabular}
% \end{table}

\noindent\textbf{Parallel Detectors.}
In addition to the ablation studies on the modules of the proposed pipeline, we also study alternatives to the structure of the pipeline. SparseFusion uses separate 3D object detectors for the LiDAR and camera modalities in parallel for extracting instance-level features to address the cross-modality domain gap. Alternatively, we consider a sequential pipeline where the camera detector runs after the LiDAR detector. The camera-modality detector inherits the output queries from the LiDAR detector. We consider two variants of this inheritance: 1) using the 3D position and instance features for the initial query of the camera modality; 2) using the 3D position but initializing the instance features from the corresponding image features. The camera detector follows the structure of the PETR~\cite{liu2022petr} decoder (one layer). Tab.~\ref{tab:ablate-parallel} shows that both sequential structures yield notably inferior performance, justifying our use of parallel modality-specific detectors.

% \vspace{-0.3cm}
\section{Conclusion}
We revisit previous LiDAR-camera fusion works and propose SparseFusion, a novel 3D object detection method that utilizes the rarely-explored strategy of fusing sparse representations. SparseFusion extracts instance-level features from each modality separately via parallel 3D object detectors and then treats the instance-level features as the modality-specific candidates. Afterward, we transform the candidates into a unified 3D space, and we are able to fuse the candidates with a lightweight attention module.
Extensive experiments demonstrate that SparseFusion achieves state-of-the-art performance on the nuScenes benchmark with the fastest inference speed. We hope SparseFusion will serve as a powerful and efficient baseline for further research into this field.

{\small
\bibliographystyle{ieee_fullname}
\bibliography{arxiv}
}

\clearpage

\begin{appendix}
In Sec.~\ref{sec:experiment}, we provide additional experimental results of SparseFusion and complementary details of the experiments presented in the main paper. Then, in Sec.~\ref{sec:network}, we elaborate on the details of the architecture of SparseFusion.

\section{Additional Experiments}
\label{sec:experiment}
\begin{figure*}[htb!]
    \centering
    \includegraphics[width=0.9\linewidth]{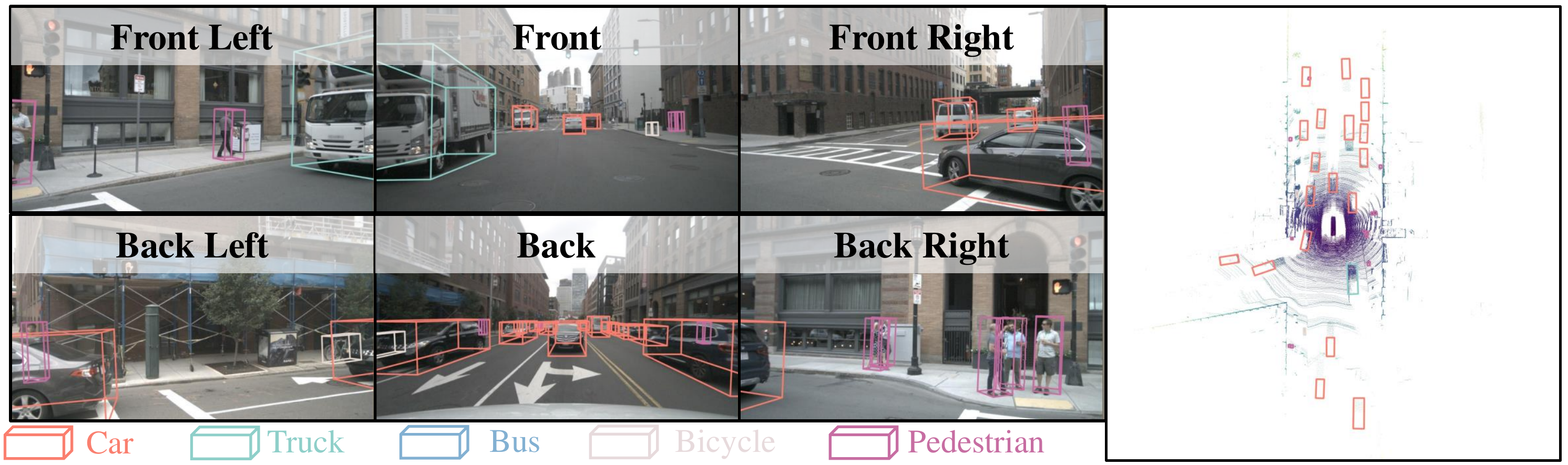}
    \includegraphics[width=0.9\linewidth]{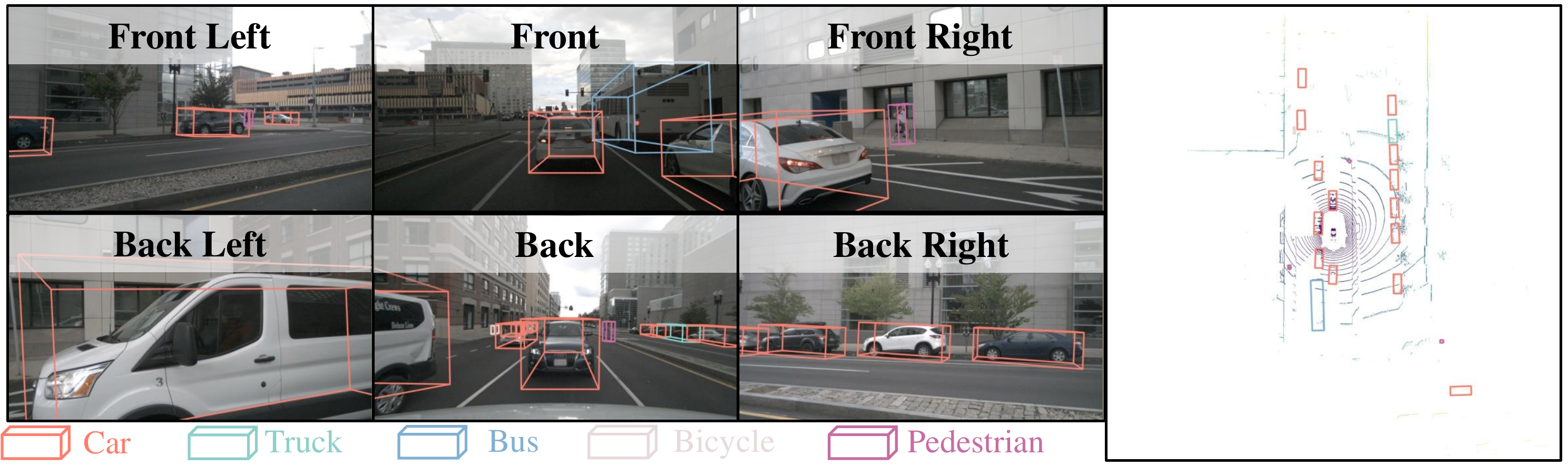}
    \caption{Qualitative results of SparseFusion on nuScenes \textit{validation set}.}
    \label{fig:visualize}
\end{figure*}

\subsection{Category-wise Results}
In Tab.~\ref{tab:category}, we report the performance of SparseFusion and our LiDAR-only baseline~\cite{bai2022transfusion} for each object category in nuScenes \textit{validation set}. SparseFusion achieves significant performance improvement for all of the object categories. In particular, the introduction of camera inputs helps to distinguish objects with similar shapes like motorcycles and bicycles.

\begin{table*}[htb!]
    \centering
    \caption{Category-wise performance on nuScenes \textit{validation set} including the overall NDS, mAP, and AP for each category.}
    \label{tab:category}
    \resizebox{\linewidth}{!}{
    \begin{tabular}{c|c|cc|cccccccccc}
        \toprule
        Methods & Modality & NDS & mAP & car & truck & bus & trailer & \makecell[c]{const.\\vehicle} & pedestrian & motorcycle & bicycle & \makecell[c]{traffic\\cone} & barrier\\
        \midrule
        TransFusion-L~\cite{bai2022transfusion} & L & 70.2 & 65.1 & 86.5 & 59.6 & 74.4 & 42.2 & 25.4 & 86.6 & 72.1 & 56.0 & 74.1 & 74.1\\
        SparseFusion & L+C & \textbf{72.8} & \textbf{70.4} & \textbf{88.5} & \textbf{64.4} & \textbf{77.1} & \textbf{44.3} & \textbf{30.3} & \textbf{89.8} & \textbf{81.5} & \textbf{71.0} & \textbf{80.6} & \textbf{76.6}\\
        \bottomrule
    \end{tabular}
    }
\end{table*}

\subsection{Qualitative Results}
We provide additional qualitative results in Fig.~\ref{fig:visualize}, where SparseFusion effectively detects most objects in the scene with the correct classification.

\subsection{Experiment Details}
\begin{figure}[h]
    \centering
    \includegraphics[width=0.48\linewidth]{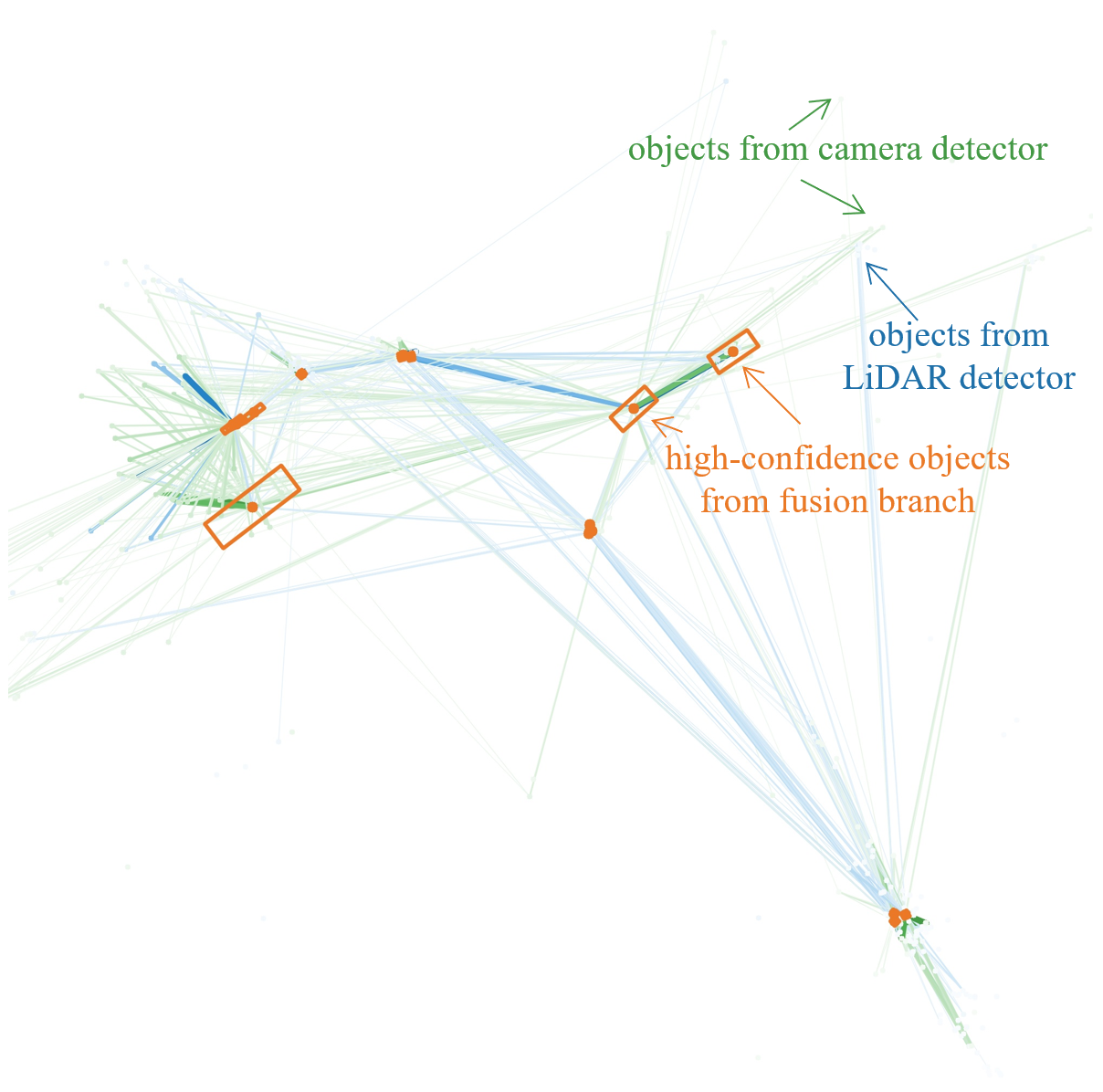}
    \hfill
    \includegraphics[width=0.48\linewidth]{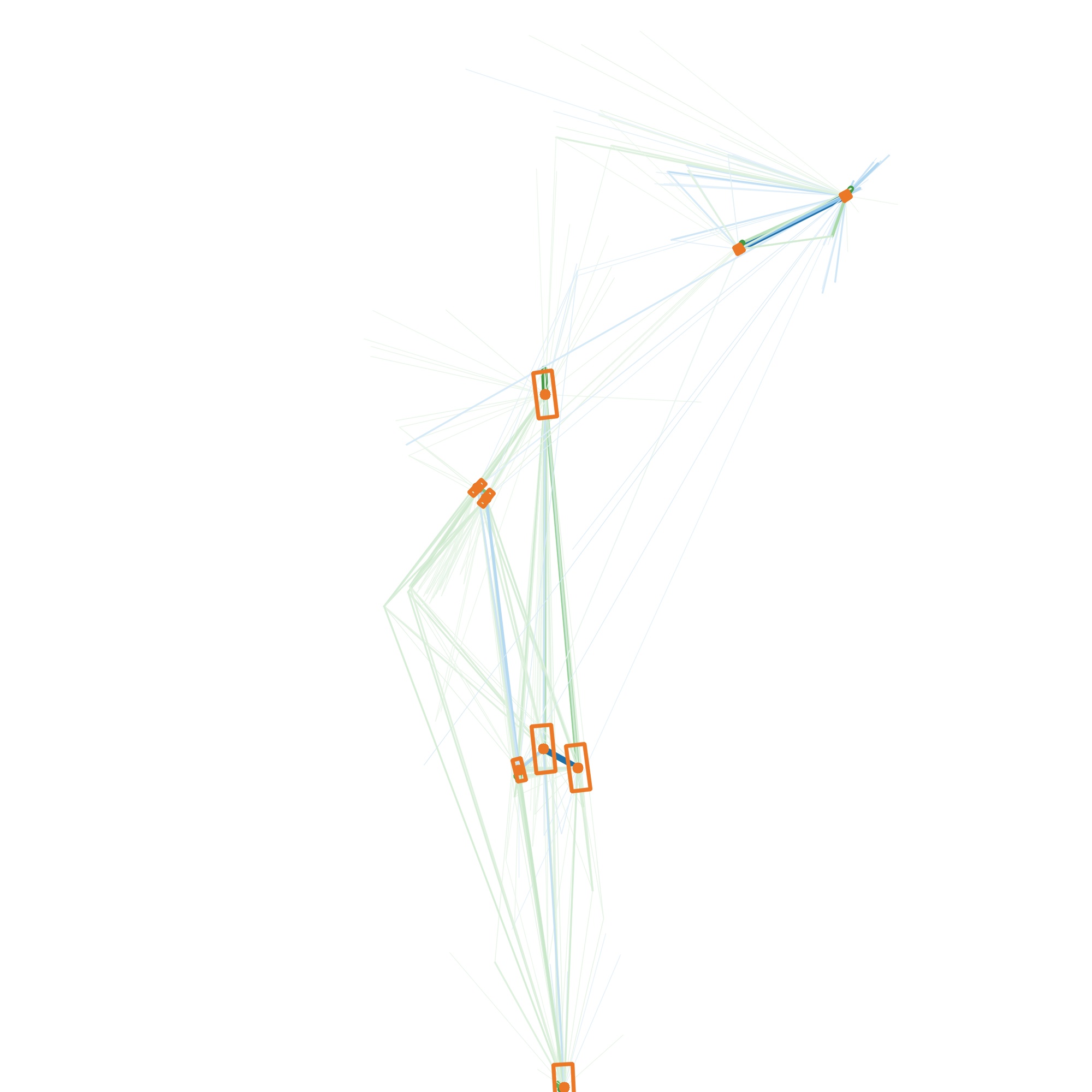}
    \includegraphics[width=0.48\linewidth]{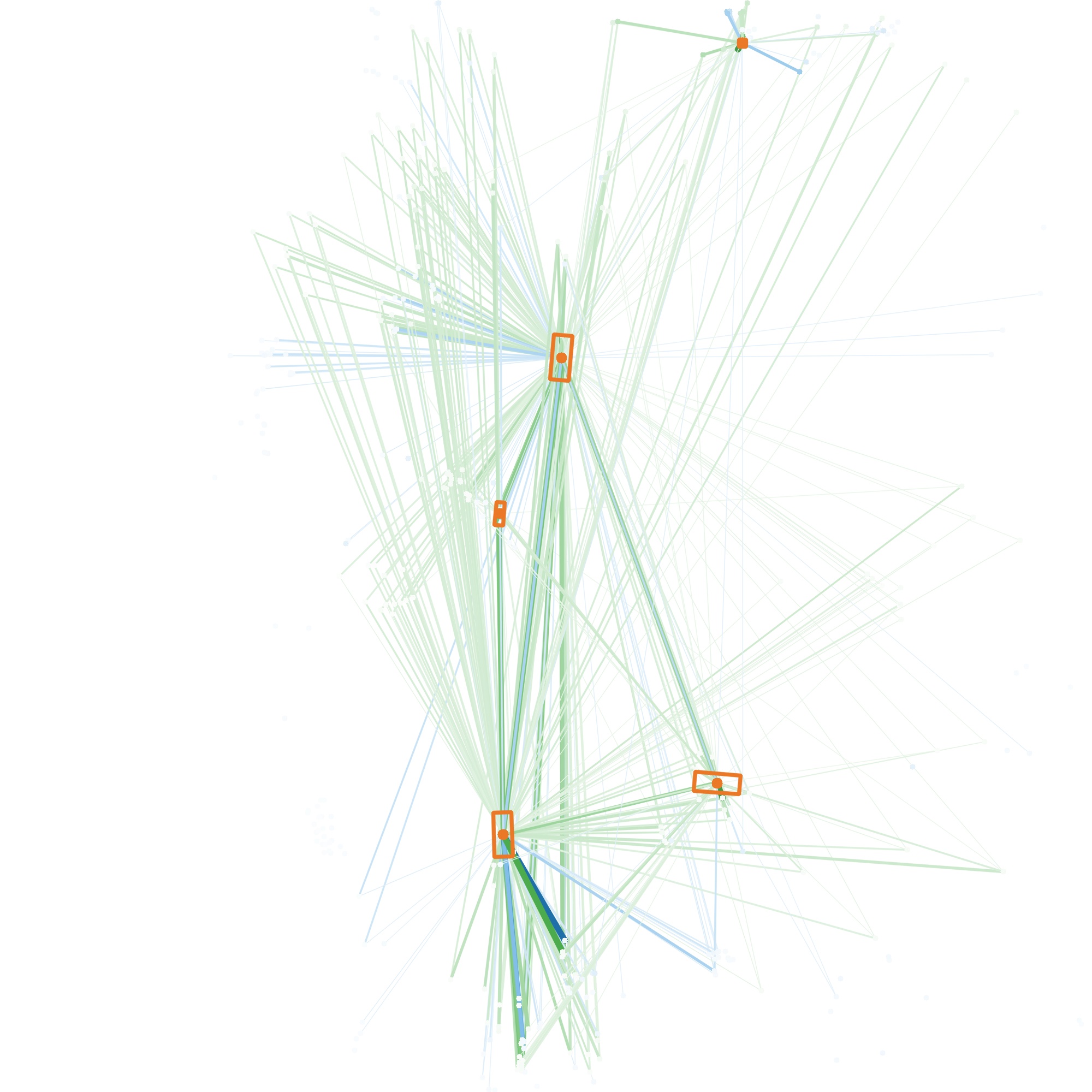}
    \hfill
    \includegraphics[width=0.48\linewidth]{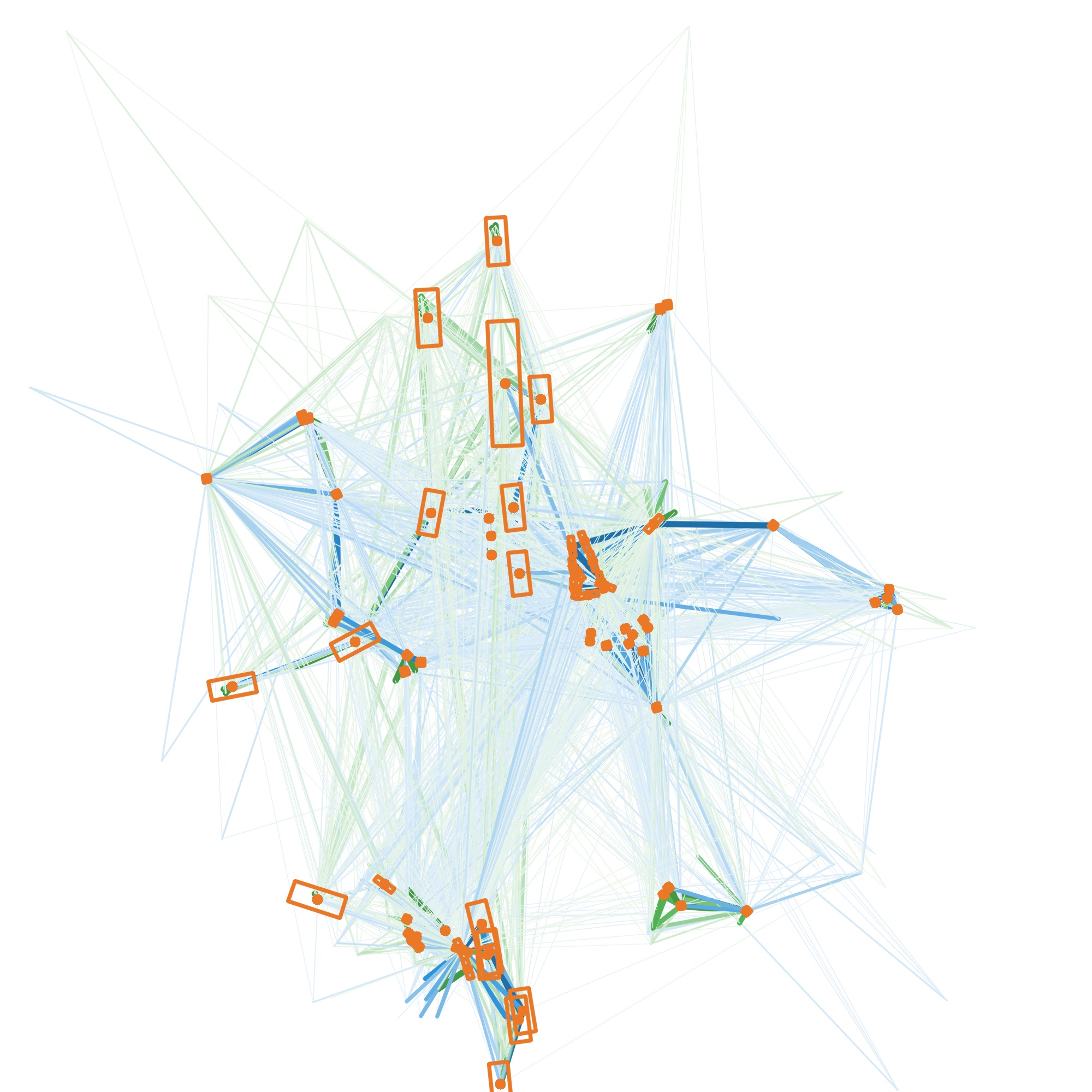}

    \caption{Instance-level feature interaction in the fusion stage. \textcolor{Orange}{Orange} boxes are objects detected after the fusion stage with high confidence in the BEV space. \textcolor{DarkBlue}{Blue} and \textcolor{BrightGreen}{green} dots denote all instances from the LiDAR and camera branches separately. \textcolor{Orange}{Orange} boxes are connected with \textcolor{DarkBlue}{blue}/\textcolor{BrightGreen}{green} dots with \textcolor{DarkBlue}{blue}/\textcolor{BrightGreen}{green} lines. The strength of attention is represented by the darkness and thickness of the lines.} % we may add two more such figures tomorrow.
    \label{fig:attention}

\end{figure}

\paragraph{Cross-Modality Sparse Representation Interaction} We provide more details about Fig.~\ref{fig:attention}. The \textcolor{Orange}{orange} boxes refer to the high-confidence objects detected by the fusion branch. The \textcolor{DarkBlue}{blue} and \textcolor{BrightGreen}{green} dots denote all instances from the LiDAR and camera branches separately even if they only have very low confidence. The \textcolor{DarkBlue}{blue}/\textcolor{BrightGreen}{green} lines separately connect the \textcolor{Orange}{orange} boxes and \textcolor{DarkBlue}{blue}/\textcolor{BrightGreen}{green} dots. We only visualize the distribution of attention for high-confidence objects detected in the fusion branches (\textcolor{Orange}{orange}). The magnitude of relationships (\textit{i.e.}, the attention value) is represented by the darkness and thickness of the lines. More examples are visualized in Fig.~\ref{fig:attention}.

\paragraph{Optimal Transport for Sparse Fusion (Tab.~\ref{tab:ablate-fusion})} We explain some details of the optimal transport strategy for sparse fusion in our ablation study. We model the distribution of LiDAR candidates as follows.
\begin{equation}
    \mathbf{p}_L(\mathbf{q}_{L,i})=\frac{s_{L,i}}{\sum_{i=1}^{N_L}s_{L,i}},i=1,2,\dots,N_L
\end{equation}
where $s_{L,i}$ is the classification confidence (highest category) of the $i$-th instance for the LiDAR detector. Similarly, the distribution of camera candidates is modeled as follows.
\begin{equation}
    \mathbf{p}_C(\mathbf{q}_{C,j})=\frac{s_{C,j}}{\sum_{i=1}^{N_C}s_{C,j}},i=1,2,\dots,N_C
\end{equation}
where $s_{C,j}$ is the classification confidence (highest category) of the $j$-th instance for the camera detector (after view transformation). We construct a cost matrix $\mathbf{C}=[c_{ij}],i=1,2,\dots,N_L,j=1,2,\dots,N_C$, where $c_{ij}$ is the euclidean distance between the centers of the $i$-th LiDAR instance and $j$-th camera instance on the BEV plane. We solve an optimal transport between $\mathbf{p}_L(\mathbf{q}_{L,i})$ and $p_C(\mathbf{q}_{C,i})$ using the IPOT algorithm~\cite{wang2019deep} which outputs an optimal transport plan $\mathbf{T}^*$, where
\begin{align}
    &\mathbf{T}^*=arg\underset{\mathbf{T}\in\mathbf{R}_+^{N_L\times N_C}}{\min}<\mathbf{C},\mathbf{T}>\\
    s.t.\ \ \  & \mathbf{T}\mathbf{1}_{N_C}=\mathbf{p}_L,\mathbf{T}^T\mathbf{1}_{N_L}=\mathbf{p}_{C}
\end{align}
We normalize $\mathbf{T}$ for each row as $\mathbf{\hat{T}}_{ij}=\mathbf{T}_{ij}/\sum_{j=1}^{N_C}\mathbf{T}_{ij}$. Then, we concatenate LiDAR candidates $\mathbf{Q}_L$ with the weighted camera candidates $\mathbf{\hat{T}}\mathbf{Q}_C$ (matrix product) in a channel-wise manner. The output features are fed into a feed-forward network to get the $N_L$ fused instance features, then the prediction head can get the object categories and bounding boxes based on the instance features.

\section{Architecture Details}
\label{sec:network}
In this section, we explain the detailed structure of each module in SparseFusion. In addition, we also illustrate the query initialization process for both LiDAR and camera detectors.

\subsection{Network Architecture}
\paragraph{LiDAR Detector} We follow TransFusion-L~\cite{bai2022transfusion} to adopt a transformer-based LiDAR detector. The initial LiDAR queries $\mathbf{Q}_L^0$ (Sec.~\ref{sec:init}) are passed through a self-attention module, then cross-attention is conducted with the BEV features from the LiDAR backbone. The output queries are fed into a feed-forward network to get the LiDAR candidates $\mathbf{Q}_L$. In both the self-attention and cross-attention modules, we add a positional encoding to all of the queries, keys, and values. Instead of the fixed sine positional embedding~\cite{vaswani2017attention}, we apply the learned embeddings by inputting the 2D XY locations of the queries, keys, and values on the BEV plane to an MLP encoder. A LiDAR view prediction head (Sec.~\ref{sec:head}) is attached to the LiDAR candidates $\mathbf{Q}_L$ to get the object category as well as the 3D bounding box in LiDAR coordinates.

\paragraph{Camera Detector} We extend Deformable-DETR~\cite{zhu2020deformable} to the 3D object detection task. The initial camera queries $\mathbf{Q}_C^0$ (Sec.~\ref{sec:init}) go through a self-attention module, then deformable attention is conducted with the image features, where we aggregate multi-scale image features from FPN~\cite{lin2017feature} through deformable attention. In deformable attention, each query only interacts with its corresponding single-view image features. The output queries are fed into a feed-forward network to get the perspective view camera candidates $\mathbf{Q}_C^P$. As we do with the LiDAR detector, we add positional embeddings to all of the queries, keys, and values, which indicate their 2D locations on the image of the corresponding view. A perspective view prediction head (Sec.~\ref{sec:head}) is attached to the perspective view camera candidates $\mathbf{Q}_C^P$ to get the object category as well as the 3D bounding box in camera coordinates.

\paragraph{View Transformation}
Our view transformation module consists of two parts: feature projection and multi-view aggregation. The feature projection is already described in Eq.~\ref{eq:view}, which encodes the camera parameters and projected boxes with two MLPs and combines them with the original instance features. The multi-view aggregation is based on a self-attention module. The output instance features belonging to all the different views are put together as $\mathbf{Q}_C^L=\{\mathbf{q}_{C,i}^L\}_{i=1}^{N_C}$. They are fed into a self-attention module and feed-forward layer. For positional embeddings added to each instance feature, we take into account both the predicted box center on the image from the camera detector and the box center on the BEV plane after bounding box coordinate transformation. The $4$-dimensional inputs are passed through an MLP to get the positional embedding for each instance feature. The updated queries serve as the camera candidates $\mathbf{Q}_C=\{\mathbf{q}_{C,i}\}_{i=1}^{N_C}$. We also attach a LiDAR view prediction head to the candidates to predict the object category and 3D bounding boxes in the LiDAR coordinates.

\paragraph{Fusion Branch}
We process the LiDAR candidates $\mathbf{Q}_L=\{\mathbf{q}_{L,i}\}_{i=1}^{N_L}$  and camera candidates $\mathbf{Q}_C=\{\mathbf{q}_{C,i}\}_{i=1}^{N_C}$ with two separate modules $f_L(\cdot),f_C(\cdot)$, each consisting of a fully-connected layer and layer normalization. Then, we concatenate the candidates as $\mathbf{Q}_{LC}=\{\mathbf{q}_{LC,i}\}_{i=1}^{N_L+N_C}$. Afterward, $\mathbf{Q}_{LC}$ is fed into a self-attention module and a feed-forward network to get the final fused instance features $\mathbf{Q}_{F}$. In the self-attention module, we also add a learned positional embedding to the instance features by encoding the XY box centers on the BEV with an MLP. Finally, we attach a LiDAR view prediction head to $\mathbf{Q}_{F}$ to predict the object category and 3D LiDAR view bounding boxes as the final results.

\begin{algorithm}[ht!]
    \caption{\textbf{Geometric Transfer}}
    \label{alg:geometry}
    \KwInput{Multi-scale image feature map of view $v$: $\mathbf{F}_{C,v}=[\mathbf{F}_{C,v}^l]_{l=0}^L$, sparse depth map of view $v$: $\mathbf{D}_{v}$.}
    \KwOutput{Multi-scale depth-aware image feature map of view $v$: $\mathbf{\hat{F}}_{C,v}=[\mathbf{\hat{F}}_{C,v}^l]_{l=0}^L$}

    $\mathbf{F}_D=\text{Stem}(\mathbf{D}_{v})$\\
    $\mathbf{\hat{F}}_{C,v}=[]$\\
    \For{$l=1,2,\dots,L$}{
        $\mathbf{F}_D=\text{Residual-Block}_l(\mathbf{F}_D)$\\
        $\mathbf{F}_D=\text{Concatenate}(\mathbf{F}_D,\mathbf{F}_{C,v}^l)$\\ \tcc{channel-wise concatenation}
        $\mathbf{F}_D=\textit{Conv}_{3\times3}^l(\mathbf{F}_D)$\\ 
        Append $\mathbf{F}_D$ to $\mathbf{\hat{F}}_{C,v}$ as $\mathbf{\hat{F}}_{C,v}^l$.
    }
    \textbf{Return} $\mathbf{\hat{F}}_{C,v}$
\end{algorithm}

\paragraph{Geometric Transfer}
We project the LiDAR point clouds to multi-view images with camera parameters to get the sparse depth maps ($200\times 112$ for nuScenes) for each view. We combine the original multi-level image features from FPN~\cite{lin2017feature} with the sparse depth map to obtain the multi-level depth-aware image features as shown in Alg.~\ref{alg:geometry}, where: $L$ is the scale level number ($L=4$ in our experiments); Stem$(\cdot)$ is a stem block composed of a $3\times 3$ convolution, batch normalization, and a ReLU activation; Residual-Block$(\cdot)$ is the basic residual block in ResNet-18~\cite{he2016deep} with stride $2$ for downsampling. Since we have multi-view images describing the surrounding scene, we run Alg.~\ref{alg:geometry} separately for each view with the shared network parameters.

\paragraph{Semantic Transfer}
Given the dense BEV features $\mathbf{F}_L\in\mathbb{R}^{H\times W\times C}$, only a few positions are indeed covered by the LiDAR point clouds. For a position $(x_j,y_j),x_j\in\{1,2,\dots,W\},y_j\in\{1,2,\dots,H\}$ on the BEV feature map occupied by point clouds, we denote the median height of the points in this pillar $(x_j,y_j)$ as $z_j$. We project all these $\{(x_j,y_j,z_j)\}$ from LiDAR coordinates to the multi-view images. We fetch these image features at these positions (max-pooling to aggregate multi-scale image features), and we combine them with the original corresponding BEV features through element-wise addition. The added features serve as the queries to interact with the multi-scale image features through a deformable-attention module and a feed-forward network. We add the positional embeddings, which are the 2D locations on the images, to the queries, keys, and values. This process is run separately for images of each view. If $(x_j,y_j,z_j)$ can be projected to multiple views, we perform max-pooling to aggregate the updated queries from multiple views. Each updated query replaces the original BEV features $\mathbf{F}_L$ at $(x_j,y_j)$ to obtain the semantic-aware BEV features $\mathbf{\hat{F}}_L$, which will be used for the query initialization of the LiDAR detector (Sec.~\ref{sec:init}).

\subsection{Prediction Head}
\label{sec:head}
We use two different prediction heads for 3D objects in the perspective view and the LiDAR view. 
\paragraph{Perspective View Head} The perspective view prediction head is designed for the camera detector to detect objects in the camera coordinates. The head includes six independent MLPs as follows:
\begin{enumerate}
    \item It predicts the category of each object. The output dimension is the number of object categories, denoting the confidence of each category.
    \item For the image of each view, it regresses the offset of the projected center of each object in the image from the reference points indicated by the positional embedding. The output dimension is two, denoting the XY coordinate separately.
    \item For the image of each view, it estimates the depths of each object. The output dimension is one.
    \item It regresses the logarithms of the XYZ scale of the 3D bounding box. The output dimension is three.
    \item It predicts the angle of each object around the vertical axis (Y-axis in the camera coordinate). The output dimension is two, denoting the $\sin$ and $\cos$ of this angle.
    \item It predicts the velocity in the horizontal plane (XZ-plane in the camera coordinate space). The output dimension is two,denoting the velocities along the X-axis and the Z-axis.
\end{enumerate}

\paragraph{LiDAR View Head} The LiDAR view prediction head is designed to detect objects in the perspective view. The same head is used for the LiDAR detector, view transformation, and the fusion branch with different network weights. The head includes six independent MLPs as follows:
\begin{enumerate}
    \item It predicts the category of each object. The output dimension is the number of object categories, denoting the confidence of each category.
    \item It regresses the offset of the center of each object on the BEV plane from the reference points indicated by the positional embedding. The output dimension is two, denoting the XY coordinate separately.
    \item It regresses the height of each object center. The output dimension is one.
    \item It regresses the logarithms of the XYZ scale of the 3D bounding box. The output dimension is three.
    \item It predicts the angle of each object around the vertical axis (Z-axis in the LiDAR coordinate space). The output dimension is two, denoting the $\sin$ and $\cos$ of this angle.
    \item It predicts the velocity in the horizontal plane (XY-plane in the LiDAR coordinate). The output dimension is two, denoting the velocities along the X-axis and the Y-axis.
\end{enumerate}

\subsection{Query Initialization}
\label{sec:init}
We follow CenterFormer~\cite{zhou2022centerformer} and TransFusion~\cite{bai2022transfusion} to initialize our queries using a heatmap, which helps to accelerate the convergence and reduce the number of queries.
\paragraph{Initialization for LiDAR Detector} 
We splatter the bounding box centers on the BEV onto a category-aware heatmap $\mathbf{Y}\in[0,1]^{H\times W\times K}$~\cite{law2018cornernet,zhou2019objects}, where $K$ is the category number, with a Gaussian kernel $\mathbf{Y}_{x,y,k_i}=\exp\left[\frac{(x-c_{x,i}^L)^2+(y-c_{y,i}^L)^2}{2\sigma_i^2}\right]$, where $k_i$ is the category of the $i$-th object, $(c_{x,i},c_{y,i})$ is its center on the BEV, and $\sigma_i$ is a standard deviation related to the object scale as done in \cite{bai2022transfusion}. The heatmap is calculated for each object separately, and we combine the multiple-object heatmaps by using the maximal value at each location. A dense head composed of $3\times 3$ convolutions are attached to the BEV features $\mathbf{\hat{F}}_L$, which is augmented by the semantic transfer. Positions $\mathbf{p}^0_{L,i}\in\mathbb{R}^2,i=1,\dots,N_L$ on the BEV with the highest confidence scores in $\max_{k_i}\mathbf{\hat{Y}}_{x_i,y_i,k_i}$ are selected as the reference points on BEV plane, along with their categories $\{k_i\}_{i=0}^{N_L}$. The local BEV features at these positions from $\mathbf{F}_L$ are fetched. We add the local feature from $\mathbf{F}_L$ and a learnable category embedding $\{\mathbf{e}^L_{k_i}\}_{i=0}^{N_L}$ to get the initial LiDAR query features $\mathbf{Q}_L^{0}=\{\mathbf{q}^{0}_{L,i}\}_{i=1}^{N_L}$.

\paragraph{Initialization for Camera Detector} For the camera modality, 3D box centers are projected into the multi-view images. We follow FCOS~\cite{tian2019fcos} to divide the objects of different sizes after projection into certain levels of multi-scale image features. We set the size thresholds to $0,48,96,192,+\infty$. For each bounding box projected on the image plane, if its $\max(length,width)$ falls between the $i$-th and $i+1$-th threshold, the object is assigned to the $i$-th scale level. As we do in the LiDAR modality, corresponding projected centers of each feature level are splattered onto a heatmap. We also get the reference points $\mathbf{p}_C^{0}=\{\mathbf{p}^{0}_{C,i}\}_{i=1}^{N_C}$ with top confidence scores from the multi-view image features, as well as the corresponding categories $\{k_i\}_{i=0}^{N_C}$. The corresponding features from the depth-aware image features $\hat{\mathbf{F}}_C$ are added with the learnable category embedding $\{\mathbf{e}^C_{k_i}\}_{i=0}^{N_C}$ to get the initial camera query features $\mathbf{Q}_C^{0}=\{\mathbf{q}^{0}_{L,i}\}_{i=1}^{N_L}$. It is worth mentioning that 3D objects are projected to multi-scale multi-view images, so initial queries come from the image features from different views and different scales.

\end{appendix}

\end{document}